\documentclass[letterpaper]{article}

\usepackage{aaai2026}

\usepackage{times}
\usepackage{helvet}
\usepackage{courier}
\usepackage[hyphens]{url}
\usepackage{graphicx}
\usepackage{natbib}
\usepackage{caption}
\usepackage{diagbox}
\usepackage{amsmath}
\usepackage{amssymb}
\usepackage{amsfonts}
\usepackage{booktabs}
\usepackage{multirow}
\usepackage{array}
\usepackage{colortbl}
\usepackage{xcolor}
\usepackage{subcaption}

\usepackage[linesnumbered,ruled,vlined]{algorithm2e}

\SetKwInput{KwInput}{Input}
\SetKwInput{KwOutput}{Output}
\SetKwRepeat{Do}{do}{while}

\SetCommentSty{mycommfont}
\DontPrintSemicolon

\newcommand{\best}[1]{\textbf{#1}}
\newcommand{\snd}[1]{\underline{#1}}
\newcommand{\trd}[1]{\textit{#1}}

\newcommand{\R}{\mathbb{R}}
\newcommand{\bA}{\mathbf{A}}
\newcommand{\bAt}{\tilde{\mathbf{A}}}
\newcommand{\bD}{\mathbf{D}}
\newcommand{\bX}{\mathbf{X}}

\newcommand{\bz}{\mathbf{z}}
\newcommand{\bh}{\mathbf{h}}

\newcommand{\bx}{\mathbf{x}}

\newcommand{\cG}{\mathcal{G}}

\newcommand{\cN}{\mathcal{N}}

\newcommand{\cM}{\mathcal{M}}
\newcommand{\cL}{\mathcal{L}}

\DeclareMathOperator*{\argmin}{arg\,min}

\title{RHEA: Reliability-Harmonized Reconstruction and Assignment\\ for Robust Multimodal-Attributed Graph Clustering}

\author{
    Yinlin Zhu\textsuperscript{1},
    Di Wu\textsuperscript{1}\thanks{Corresponding author.},
    Ziyu Han\textsuperscript{2},
    Zekai Chen\textsuperscript{3},\\
    Wang Luo\textsuperscript{1},
    Miao Hu\textsuperscript{1},
    Guocong Quan\textsuperscript{1}
}
\affiliations{

    Sun Yat-sen University, Guangzhou, China\textsuperscript{\rm 1}\\
    Shandong University, Weihai, China\textsuperscript{\rm 2}\\
    Beijing Institute of Technology, Beijing, China\textsuperscript{\rm 3}\\

    \{zhuylin27,\ luow69\}@mail2.sysu.edu.cn, 
    \{wudi27,\ humiao5,\ quangc\}@mail.sysu.edu.cn, \\
    hanziyu@mail.sdu.edu.cn, zackchen02@163.com
}

\begin{document}
\maketitle

\begin{abstract}
Multimodal-attributed graphs (MAGs), whose nodes carry heterogeneous attributes such as text and images over a relational structure, have become a fundamental substrate for label-free entity grouping tasks, including community discovery and product segmentation. Existing MAG clustering methods effectively integrate complementary modalities when attributes are clean and complete, but degrade substantially under noisy or missing attributes because they implicitly assume equal modality reliability across all nodes. In practice, modality reliability is inherently node-specific: images may be corrupted or absent, while textual descriptions are incomplete or noisy. We argue that, under attribute homophily, graph neighborhoods naturally provide supervision-free evidence for estimating node-specific modality reliability. Based on this insight, we propose RHEA, a reliability-aware framework for MAG clustering that estimates node-specific modality reliability from neighborhood consensus and propagates this signal throughout the clustering pipeline. RHEA reconstructs unreliable or missing modalities from graph neighborhoods, adaptively weights modalities during reliability-aware fusion, and performs topology-aware optimal transport clustering with reliability-aware transport assignment and neighbor-consensus assignment distillation. Furthermore, the confidence of reconstructed representations is incorporated into the clustering objective, allowing uncertain reconstructions to contribute proportionally during optimization. Experiments on four MAG benchmarks under five attribute conditions show that RHEA consistently outperforms the strongest baseline, with NMI gains increasing as attribute quality deteriorates.
\end{abstract}

\section{Introduction}

Multimodal-attributed graphs (MAGs), in which nodes are associated with heterogeneous attributes such as text and images over relational structures, have become a fundamental data model for recommendation systems, e-commerce catalogs, citation networks, and social communities~\citep{guo2025dmgc,zheng2025dgf,peng2021agcn}. A fundamental task on MAGs is node clustering, which aims to discover semantically coherent node clusters without manual annotation for applications such as community discovery, product segmentation, and cold-start taxonomy construction. Recent multimodal graph clustering methods learn node representations by jointly modeling graph topology and multimodal attributes, achieving strong performance when attributes across all modalities are clean and complete~\citep{guo2025dmgc,zheng2025dgf}.

However, this assumption rarely holds in real-world MAGs. Product images may be corrupted or missing, textual descriptions are often noisy or incomplete, and the most informative modality can vary substantially across nodes. Existing methods~\citep{guo2025dmgc,zheng2025dgf} nevertheless implicitly assume that each modality is equally reliable for all nodes during representation learning. As a result, corrupted modalities contaminate fused representations, missing modalities cannot be effectively compensated, and clustering performance deteriorates as attribute quality declines. The problem is particularly challenging in the unsupervised setting, where modality reliability cannot be inferred from labels. This raises a fundamental question: \emph{how can node-specific modality reliability be estimated without supervision to support robust multimodal-attributed graph clustering?}

We argue that the graph structure itself provides the answer. Under the widely observed principle of attribute homophily~\citep{mcpherson2001homophily}, neighboring nodes tend to exhibit similar semantic attributes. Therefore, a modality that is consistent with its neighborhood is more likely to be reliable than one that persistently deviates from local consensus. This observation enables supervision-free estimation of node-specific modality reliability directly from the graph. Instead of treating reliability as a modality-level property shared across all nodes, we model it as a latent variable that varies across both nodes and modalities, providing a unified signal for robust multimodal representation learning.

Building upon this insight, we propose \underline{\textbf{R}}eliability-\underline{\textbf{H}}armonized r\underline{\textbf{E}}construction and \underline{\textbf{A}}ssignment (\textbf{RHEA}), a reliability-aware framework for MAG clustering. RHEA estimates node-specific modality reliability from neighborhood consensus and leverages it to reconstruct unreliable modalities, adaptively fuse multimodal representations, and guide topology-aware optimal transport clustering. A unified reliability signal connects these components, enabling robust representation learning under both attribute corruption and missingness. Extensive experiments on four benchmark datasets under five attribute conditions show that RHEA consistently outperforms existing methods, with improvements becoming larger as attribute quality deteriorates.
Moreover, the learned reliability estimates accurately recover synthetically injected attribute corruption, with AUROC consistently exceeding $0.95$ across all datasets, validating the effectiveness of the proposed reliability modeling.

\textbf{Our contributions.} (1) \textbf{Valuable Insights.} We identify \emph{node-specific modality reliability} as the missing ingredient for robust multimodal-attributed graph clustering, and show that graph neighborhoods provide supervision-free evidence for estimating this reliability. (2) \textbf{Novel Method.} We propose RHEA, a unified reliability-aware framework in which a shared neighborhood-consensus reliability signal simultaneously guides graph-based modality reconstruction, adaptive multimodal fusion, and topology-aware optimal transport clustering. (3) \textbf{State-of-the-Art Performance.} Extensive experiments on four benchmark datasets demonstrate consistent improvements over state-of-the-art methods across diverse corruption and missingness settings.

\section{Related Work}

\paragraph{Attributed Graph Clustering.}
Attributed graph clustering has been extensively studied for graphs with a single attribute modality. Early methods couple graph autoencoders with clustering objectives, progressively improving structural-feature fusion through dual-branch architectures, attention mechanisms, or refined graph encoders~\citep{bo2020sdcn,tu2021dfcn,peng2021agcn,wang2019daegc,liu2022dcrn,cui2020age,liu2023dinknet,liu2022sublime}. More recent approaches instead rely on self-supervised objectives, including contrastive learning~\citep{zhu2021gca,you2020graphcl,xia2022progcl,liu2023hsan,yang2023ccgc,liu2024ns4gc} and differentiable graph partitioning based on modularity optimization~\citep{tsitsulin2023dmon,newman2006modularity,liu2024magi}. Extending clustering to multimodal attributed graphs remains comparatively underexplored. Existing methods mainly improve cross-modal representation learning through disentanglement~\citep{guo2025dmgc}, graph filtering and contrastive alignment~\citep{zheng2025dgf}, feature fusion~\citep{lin2022diagc,pan2021mcgc,ke2023cloven}, or more expressive multimodal architectures~\citep{zhu2025moegcl,li2025trilearn,li2025multiscale,li2026lion,lin2022unicross}. Despite their differences, these methods employ a shared fusion strategy across all nodes, implicitly assuming that every modality contributes equally regardless of local modality quality in real-world scenarios.

\paragraph{Incomplete Multimodal Learning.}
Our work is closely related to learning under missing or unreliable modalities. Existing incomplete multi-view methods typically recover missing views through contrastive prediction, feature alignment, or generative completion~\citep{lin2021completer,xu2022apadc}, while others adaptively weight modalities using evidential learning, information bottlenecks, or optimal transport~\citep{han2022trustedmvc,yan2024dib,cuturi2013sinkhorn,peyre2019computational,xue2025protocol}. On graphs, recent approaches extend these ideas via graph propagation~\citep{malitesta2026trainingfree,roh2025mmgf}, graph denoising~\citep{zhou2023freedom}, diffusion-based modality generation~\citep{jiang2024diffmm}, or uncertainty-aware aggregation~\citep{shim2019robust,wang2025modality,chen2025rseamvgnn}. In contrast, we neither train a dedicated generative completion model nor assume a globally shared, modality-level reliability estimate. Instead, we study multimodal graph clustering under heterogeneous modality quality, where reliability is node-specific and unreliable modalities are repaired directly from graph neighborhoods. Our evaluation follows the standardized protocols of MAGB and OpenMAG~\citep{yan2024magb,wan2026openmag}; broader discussions of incomplete graph learning can be found in surveys~\citep{xia2025incomplete,wu2024missingmodality}.

\section{Preliminaries and Problem Formulation}

\paragraph{Multimodal Attributed Graph (MAG).}
We consider a MAG $\cG=(\bA,\{\bX^{(m)}\}_{m\in\cM})$ over $N$ nodes. Here
$\bA\in\{0,1\}^{N\times N}$ is the adjacency matrix and $\cM$ indexes the attribute modalities, in our case text
and image ($|\cM|=2$). Each modality is associated with an attribute matrix $\bX^{(m)}\in\R^{N\times d_m}$ whose
row $\bx^{(m)}_i$ holds the modality-$m$ attributes of node $i$, obtained from standard pretrained
encoders~\citep{vaswani2017attention,radford2021clip}. We use $\bD$ to denote the degree matrix,
$\bAt=\bD^{-1/2}\bA\bD^{-1/2}$ for the symmetric-normalized adjacency, and $\cN(i)$ for the graph neighbors of
node $i$.

\paragraph{MAG Clustering.}
We consider MAG clustering under heterogeneous modality quality, where, for an unknown subset
of nodes, a modality is either absent (its absence is observable) or present but uninformative (its
unreliability is latent and must be inferred). The goal is to partition the $N$ nodes into $K$ clusters without
labels, producing a hard assignment $\hat{y}_i\in\{1,\dots,K\}$ for every node $i$.

\paragraph{Reliability Field.}
We define a node-specific \emph{modality reliability field} $\rho_{i,m}\in[0,1]$ that captures how
trustworthy modality $m$ is for node $i$, normalized per node so that $\sum_{m\in\cM}\rho_{i,m}=1$. Notably, a larger value
indicates higher reliability.

\section{Methodology}

\begin{figure*}[t]
\centering
\includegraphics[width=0.995\textwidth]{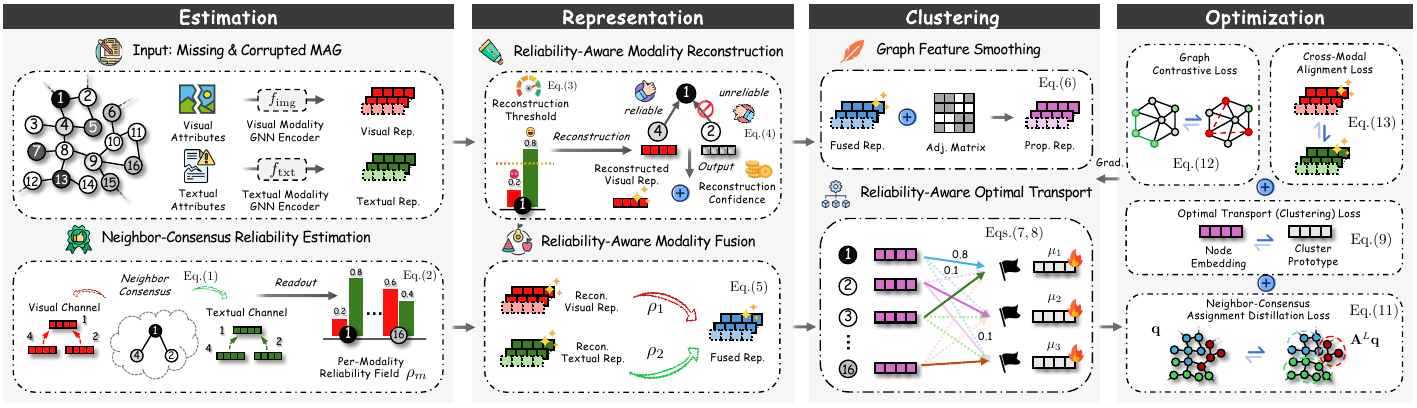}
\caption{Overview of RHEA. RHEA consists of four stages: (1) \textbf{Estimation}: learning node-specific modality reliability from neighbor consensus; (2) \textbf{Representation}: reconstructing unreliable modalities and performing reliability-aware fusion; (3) \textbf{Clustering}: obtaining topology-consistent assignments through reliability-aware optimal transport; and (4) \textbf{Optimization}: jointly training clustering and representation objectives. A unified reliability field connects all stages.}
\label{fig:framework}
\end{figure*}

In this section, we present RHEA, a reliability-aware framework for robust multimodal graph clustering.
As illustrated in Fig.~\ref{fig:framework}, RHEA estimates node-specific modality reliability from neighborhood consensus in a fully unsupervised manner, and propagates this reliability field throughout representation learning, fusion, and clustering.

\subsection{Neighbor-Consensus Reliability Estimation}
\label{sec:rho}

RHEA is built upon the observation that modality reliability is inherently node-specific. Rather than assuming that each modality is equally reliable across all nodes, we estimate a node-specific modality reliability field based on neighborhood consensus. Under attribute homophily, reliable modalities should be consistent with the semantic evidence provided by neighboring nodes, whereas corrupted modalities tend to deviate from this consensus.

Specifically, let
$\bz^{(m)}_i=f_m(\bX^{(m)},\bA)\in\R^{d}$ be the $\ell_2$-normalized embedding of node $i$ in modality $m$ from
the per-modality graph encoder $f_m$ (e.g., GCN~\citep{kipf2017gcn} or GAT~\citep{velickovic2018gat}). For node $i$ and modality $m$, we measure the intra-modal neighbor agreement, which can be calculated as follows:
\begin{equation}
a_{i,m} \;=\; \frac{1}{|\cN(i)|}\sum_{j\in\cN(i)} \big\langle \bz^{(m)}_i,\, \bz^{(m)}_j \big\rangle,
\label{eq:agree}
\end{equation}
where $\langle\cdot,\cdot\rangle$ is cosine similarity on the $\ell_2$-normalized embeddings. The agreement score measures how consistently modality $m$ of node $i$ aligns with its local neighborhood under structural homophily.

Subsequently, we convert the per-modality
agreements into a per-node reliability, which is formulated as:
\begin{equation}
\rho_{i,m} \;=\; \frac{\exp(a_{i,m}/\tau)}{\sum_{m'\in\cM}\exp(a_{i,m'}/\tau)},
\label{eq:rho}
\end{equation}
where $\tau$ is a temperature, $\rho_{i,m}\in[0,1]$ and $\sum_m\rho_{i,m}=1$. The resulting $\rho_{i,m}$ defines a probability distribution over modalities for each node, which serves as the unified reliability field for downstream learning.

\subsection{Reliability-Aware Representation Learning}
\label{sec:recon}

The estimated reliability identifies modalities that cannot be reliably used for representation learning. RHEA first reconstructs unreliable modalities from their graph neighborhoods, allowing neighborhood consensus to compensate for missing or corrupted observations before multimodal fusion.

Without loss of generality, we describe the reconstruction procedure for the image modality; the same process is applied symmetrically to all modalities (Appendix~\ref{app:modality}). Reconstruction is triggered under two conditions: (1) the modality is missing, or (2) the modality is observed but assigned low reliability. Accordingly, the reconstruction set is defined as:
\begin{equation}
\mathcal{R} = \{ i : o_i = 0 \;\vee\; \rho_{i,\mathrm{img}} < \gamma \},
\label{eq:gate}
\end{equation}
where $o_i\in\{0,1\}$ indicates whether the image modality of node $i$ is observed and $\gamma$ denotes the reliability threshold.

For every node $i\in\mathcal{R}$, the unreliable image representation is reconstructed by aggregating image embeddings from neighboring nodes with available observations:
\begin{equation}
\hat{\bz}^{(\mathrm{img})}_i
=
\frac{
\sum_{j\in\cN^{+}(i)}
\bAt_{ij}\,
\bz^{(\mathrm{img})}_j
}{
\sum_{j\in\cN^{+}(i)}
\bAt_{ij}
},
\qquad
i\in\mathcal{R},
\label{eq:recon}
\end{equation}
where
$\cN^{+}(i)=\{j\in\cN(i): o_j=1\}$
contains only neighbors with observed image modalities. Restricting message passing to observed neighbors prevents missing or corrupted representations from propagating through the graph while preserving the neighborhood consensus.

The reliability of each reconstruction further depends on the quality of its supporting neighbors. We therefore estimate a reconstruction confidence
$c_i=\exp(-\beta v_i)\cdot\frac{w_i}{w_i+1}$,
where $\beta>0$ is a decay coefficient,
$w_i=\sum_{j\in\cN^{+}(i)}\bAt_{ij}$
measures the available neighborhood evidence, and
$v_i$
is the mean per-dimension variance of the donor embeddings. Consequently, reconstruction confidence increases when more neighboring observations are available and their representations are mutually consistent, and decreases otherwise. This confidence serves as an absolute reliability measure that complements the relative reliability and is subsequently incorporated into the topology-aware clustering objective (Sec.~\ref{sec:ot}).

Finally, the reconstructed modality representations are fused according to the estimated modality reliability:
\begin{equation}
\bz_i
=
\rho_{i,\mathrm{txt}}
\bz^{(\mathrm{txt})}_i
+
\rho_{i,\mathrm{img}}
\tilde{\bz}^{(\mathrm{img})}_i,
\label{eq:fuse}
\end{equation}
where $\tilde{\bz}^{(\mathrm{img})}_i$ is the image representation used for downstream fusion.
Specifically, when node $i$ is identified as unreliable or missing ($i \in \mathcal{R}$), its image feature is substituted by the reconstructed embedding $\hat{\bz}^{(\mathrm{img})}_i$;
otherwise, the original observed embedding $\bz^{(\mathrm{img})}_i$ is retained.

\subsection{Reliability-Aware Clustering}
\label{sec:ot}

At the core of RHEA is a reliability-aware clustering mechanism that turns the
estimated modality reliability into the actual cluster assignments. It comprises two
coupled components: a reliability-aware optimal transport assignment that lets
reliable modalities dominate the transport cost (Eq.~\eqref{eq:ot}), and a neighbor-consensus assignment distillation objective that repairs assignments
where neighborhood evidence is weak (Eq.~\eqref{eq:ncrc}). Together they constitute the
primary learning signal of our model.

\paragraph{Reliability-aware Optimal Transport.}
The estimated reliability is incorporated into cluster assignment. Since multimodal
representations have already been reconstructed and fused according to their estimated
reliability (Sec.~\ref{sec:recon}), we first smooth the fused node embeddings over the
graph to integrate local topological context before clustering,
$\bh_i=\mathrm{norm}(\sum\nolimits_j\bAt_{ij}\bz_j)$,
where $\mathrm{norm}(\cdot)$ is $\ell_2$ normalization.
Cluster assignments are then obtained by solving an entropy-regularized optimal transport
problem between node embeddings and $K$ learnable cluster prototypes
$\{\boldsymbol{\mu}_k\}$. The transport cost is defined as follows:
\begin{equation}
\mathbf C_{ik}
=
1-\langle\bh_i,\boldsymbol{\mu}_k\rangle,
\label{eq:cost}
\end{equation}
where $\bh_i$ already encodes the node-specific modality reliability through
Eq.~\eqref{eq:fuse}. Consequently, unreliable modalities contribute less to the transport
cost and exert reduced influence on the resulting cluster assignment. The optimal
transport plan is then computed as:
\begin{equation}
\mathbf P^\star
=
\argmin_{\mathbf P\in\Pi(\mathbf r,\mathbf c)}
\;
\langle\mathbf P,\mathbf C\rangle
-
\epsilon H(\mathbf P),
\label{eq:ot}
\end{equation}
where $H(\cdot)$ denotes the entropy regularizer, $\epsilon$ is the regularization
strength, and $\Pi(\mathbf r,\mathbf c)$ denotes the transport polytope with row and column
marginals. Crucially, both marginals are made reliability-aware. To account for
reconstruction uncertainty, the row marginal is weighted by the reconstruction confidence
(i.e., $r_i\propto c_i$) such that nodes reconstructed from limited or inconsistent
neighborhood evidence contribute less transport mass; the column marginal $\mathbf c$ is
estimated from the graph topology to reflect the empirical cluster-size prior rather than
assuming uniform cluster proportions. The optimization problem in Eq.~\eqref{eq:ot} can be solved using entropic Sinkhorn iterations~\citep{cuturi2013sinkhorn}.

The clustering objective is then defined as follows:
\begin{equation}
    \mathcal L_{\mathrm{clu}} = \langle\mathbf P^\star,\mathbf C\rangle,
\end{equation}
where $\mathbf P^\star$ is treated as a fixed target via stop-gradient, and the final
cluster assignment is obtained as $\hat y_i=\arg\max_kP^\star_{ik}$.

\paragraph{Neighbor-consensus Assignment Distillation (NCRC).}
Although reliability-aware optimal transport substantially improves assignment robustness,
reconstructed nodes may still receive inaccurate assignments when neighborhood evidence is
insufficient. We therefore introduce a neighbor-consensus assignment distillation objective
that further regularizes cluster assignments using neighborhood consensus. Specifically, let
$
\mathbf q_i
=
\mathrm{softmax}
\!\left(
\langle\bh_i,\boldsymbol{\mu}\rangle/\tau_a
\right)
$
denote the soft cluster assignment, with assignment temperature $\tau_a$. The assignments are propagated over the graph for $L$
hops,
$\tilde{\mathbf q}=\bAt^{L}\mathbf q$,
and sharpened to a neighbor-consensus target
$\tilde q^{\mathrm{sharp}}_{ik}\propto\tilde q_{ik}^{1/T_s}$ ($T_s{<}1$, renormalized per node).
Each node is then distilled toward this consensus via:
\begin{equation}
\mathcal L_{\mathrm{ncrc}}
=
-\frac1N
\sum_i
\sum_k
\mathrm{sg}
\!\left[
\tilde q^{\mathrm{sharp}}_{ik}
\right]
\log q_{ik},
\label{eq:ncrc}
\end{equation}
where $\mathrm{sg}[\cdot]$ denotes the stop-gradient operator. This objective encourages
locally consistent cluster assignments while preventing the propagated consensus target
from being updated by the student predictions.
\subsection{Optimization Objectives}
\label{sec:contrast}

The clustering objectives introduced above provide the primary supervision signal. 
Since reliability estimation depends on encoder representations, we further regularize the encoder with two lightweight auxiliary contrastive objectives defined on the fused embeddings $\bh$. These objectives improve representation quality without directly affecting cluster assignment.

\paragraph{Auxiliary contrastive regularization.}
We first introduce a graph-walk-based contrastive loss $\cL_{\mathrm{nbr}}$, which preserves topology-aware invariance by enforcing agreement between random-walk co-occurrence pairs. 
Given an anchor node $i$, its positive set $\mathcal{P}_i$ and negative set $\mathcal{Q}_i$ are constructed from random walks (walk length 5, 10 walks per node, context window 3). 
We then optimize an InfoNCE objective with temperature $\tau_c$ over the fused representation $\bh$:
\begin{equation}
\cL_{\mathrm{nbr}} =
-\frac{1}{N}
\sum_i
\frac{1}{|\mathcal{P}_i|}
\sum_{j\in\mathcal{P}_i}
\log
\frac{
e^{\langle \bh_i,\bh_j\rangle/\tau_c}
}{
\sum_{k\in\mathcal{P}_i\cup\mathcal{Q}_i}
e^{\langle \bh_i,\bh_k\rangle/\tau_c}
}.
\label{eq:lnbr}
\end{equation}

We further introduce a cross-modal alignment loss $\cL_{\mathrm{mod}}$, which aligns modality-specific representations with the fused embedding after reliability-aware fusion. 
Let $\bz^{(\mathrm{txt})}_i$ and $\tilde{\bz}^{(\mathrm{img})}_i$ denote the text and (possibly reconstructed) image representations, respectively. The alignment objective is defined as follows:
\begin{equation}
\cL_{\mathrm{mod}} = \frac{1}{N}\sum_i
\big[
2-\langle \bh_i,\bz^{(\mathrm{txt})}_i\rangle
-\langle \bh_i,\tilde{\bz}^{(\mathrm{img})}_i\rangle
\big] + \Omega_{\mathrm{cm}},
\label{eq:lmod}
\end{equation}
where $\Omega_{\mathrm{cm}}$ is a small inter-node repulsion term that discourages trivial collapse by enforcing margin separation across nodes (Detailed in Appendix~\ref{app:complexity}).

\paragraph{Joint objective.}
The model is trained end-to-end by minimizing the following optimization objective:
\begin{equation}
\cL =
\cL_{\mathrm{clu}}
+ \lambda_{\mathrm{ncrc}}\cL_{\mathrm{ncrc}}
+ \lambda_{\mathrm{nbr}}\cL_{\mathrm{nbr}}
+ \lambda_{\mathrm{mod}}\cL_{\mathrm{mod}}.
\label{eq:joint}
\end{equation}
The first two terms constitute the core clustering objective, combining reliability-aware optimal transport assignment ($\cL_{\mathrm{clu}}$) and neighbor-consensus assignment distillation ($\cL_{\mathrm{ncrc}}$). The remaining terms act as auxiliary representation regularizers that improve encoder robustness, with $\lambda_{\mathrm{nbr}}$ and $\lambda_{\mathrm{mod}}$ controlling their influence, while $\lambda_{\mathrm{ncrc}}$ balances the two clustering losses. Algorithm~\ref{alg:rhea} summarizes the procedure.

\begin{table*}[t]
\centering
\setlength{\tabcolsep}{3pt}
\renewcommand{\arraystretch}{1.1}
\resizebox{\textwidth}{!}{
\begin{tabular}{l cccc cccc  cccc cccc}
\toprule[1pt]

\multirow{2}{*}{\textbf{Methods}} & \multicolumn{4}{c}{\textbf{RedditS}} & \multicolumn{4}{c}{\textbf{Toys}} & \multicolumn{4}{c}{\textbf{Grocery}} & \multicolumn{4}{c}{\textbf{Amazon}} \\
\cmidrule(lr){2-5}\cmidrule(lr){6-9}\cmidrule(lr){10-13}\cmidrule(lr){14-17}
 & \textbf{NMI} & \textbf{ACC} & \textbf{ARI} & \textbf{F1} & \textbf{NMI} & \textbf{ACC} & \textbf{ARI} & \textbf{F1} & \textbf{NMI} & \textbf{ACC} & \textbf{ARI} & \textbf{F1} & \textbf{NMI} & \textbf{ACC} & \textbf{ARI} & \textbf{F1} \\
\midrule[0.1pt]
KMeans &
.774$_{\pm.004}$ & .737$_{\pm.010}$ & .689$_{\pm.017}$ & .651$_{\pm.006}$ &
.308$_{\pm.004}$ & .353$_{\pm.004}$ & .171$_{\pm.001}$ & .315$_{\pm.003}$ &
.202$_{\pm.006}$ & .248$_{\pm.018}$ & .099$_{\pm.010}$ & .219$_{\pm.013}$ &
\trd{.358$_{\pm.003}$} & \snd{.799$_{\pm.001}$} & \snd{.465$_{\pm.003}$} & \trd{.716$_{\pm.002}$} \\
\midrule[0.1pt]

DFCN &
.804$_{\pm.003}$ &
\snd{.812$_{\pm.011}$} &
.765$_{\pm.013}$ &
\snd{.725$_{\pm.006}$} &
.430$_{\pm.008}$ &
\snd{.441$_{\pm.009}$} &
.255$_{\pm.009}$ &
\snd{.400$_{\pm.014}$} &
.402$_{\pm.008}$ &
.406$_{\pm.013}$ &
.255$_{\pm.011}$ &
.328$_{\pm.012}$ &
\snd{.381$_{\pm.009}$} &
\trd{.763$_{\pm.010}$} &
\trd{.427$_{\pm.019}$} &
\snd{.718$_{\pm.010}$} \\

DMoN &
.595$_{\pm.134}$ & .531$_{\pm.182}$ & .451$_{\pm.185}$ & .394$_{\pm.214}$ &
.130$_{\pm.108}$ & .177$_{\pm.067}$ & .058$_{\pm.057}$ & .103$_{\pm.097}$ &
.124$_{\pm.098}$ & .193$_{\pm.043}$ & .052$_{\pm.054}$ & .106$_{\pm.081}$ &
.347$_{\pm.051}$ & .710$_{\pm.096}$ & .363$_{\pm.112}$ & .645$_{\pm.089}$ \\

MVGRL &
.792$_{\pm.006}$ & .781$_{\pm.012}$ & .742$_{\pm.015}$ & \trd{.701$_{\pm.009}$} &
.402$_{\pm.010}$ & .415$_{\pm.012}$ & .236$_{\pm.011}$ & \trd{.382$_{\pm.010}$} &
.386$_{\pm.009}$ & .392$_{\pm.011}$ & .241$_{\pm.010}$ & .318$_{\pm.012}$ &
.333$_{\pm.010}$ & .725$_{\pm.012}$ & .401$_{\pm.013}$ & .690$_{\pm.010}$ \\

S3GC &
.702$_{\pm.012}$ & .651$_{\pm.020}$ & .602$_{\pm.018}$ & .541$_{\pm.022}$ &
.310$_{\pm.014}$ & .335$_{\pm.015}$ & .189$_{\pm.013}$ & .282$_{\pm.012}$ &
.318$_{\pm.010}$ & .341$_{\pm.012}$ & .176$_{\pm.011}$ & .264$_{\pm.013}$ &
.301$_{\pm.012}$ & .701$_{\pm.015}$ & .328$_{\pm.014}$ & .612$_{\pm.013}$ \\

\midrule[0.1pt]

COMPLETER &
.643$_{\pm.013}$ & .534$_{\pm.026}$ & .476$_{\pm.033}$ & .400$_{\pm.030}$ &
.343$_{\pm.016}$ & .340$_{\pm.013}$ & .188$_{\pm.014}$ & .277$_{\pm.011}$ &
.370$_{\pm.033}$ & .381$_{\pm.025}$ & .218$_{\pm.036}$ & .297$_{\pm.022}$ &
.189$_{\pm.089}$ & .569$_{\pm.156}$ & .109$_{\pm.201}$ & .462$_{\pm.148}$ \\

APADC &
.540$_{\pm.056}$ & .339$_{\pm.086}$ & .239$_{\pm.059}$ & .227$_{\pm.073}$ &
.262$_{\pm.018}$ & .246$_{\pm.019}$ & .071$_{\pm.016}$ & .179$_{\pm.023}$ &
.283$_{\pm.028}$ & .285$_{\pm.033}$ & .126$_{\pm.026}$ & .164$_{\pm.022}$ &
.064$_{\pm.025}$ & .649$_{\pm.013}$ & .045$_{\pm.023}$ & .330$_{\pm.034}$ \\

\midrule[0.1pt]

DGF &
\snd{.836$_{\pm.003}$} &
\trd{.807$_{\pm.014}$} &
\snd{.811$_{\pm.013}$} &
.695$_{\pm.031}$ &
\snd{.434$_{\pm.011}$} &
\trd{.424$_{\pm.018}$} &
\snd{.268$_{\pm.011}$} &
.372$_{\pm.015}$ &
\snd{.456$_{\pm.003}$} &
\snd{.411$_{\pm.005}$} &
\snd{.295$_{\pm.003}$} &
\snd{.348$_{\pm.008}$} &
.341$_{\pm.011}$ &
.740$_{\pm.013}$ &
.372$_{\pm.020}$ &
.673$_{\pm.011}$ \\

DMGC &
\trd{.821$_{\pm.005}$} &
.799$_{\pm.013}$ &
\trd{.789$_{\pm.011}$} &
.681$_{\pm.021}$ &
\trd{.432$_{\pm.010}$} &
.417$_{\pm.015}$ &
\trd{.262$_{\pm.012}$} &
.364$_{\pm.013}$ &
\trd{.438$_{\pm.005}$} &
\trd{.408$_{\pm.006}$} &
\trd{.283$_{\pm.004}$} &
\trd{.340$_{\pm.008}$} &
.352$_{\pm.008}$ &
.749$_{\pm.010}$ &
.389$_{\pm.009}$ &
.686$_{\pm.009}$ \\

\midrule[0.1pt]

RHEA (ours) &
\best{.868$_{\pm.008}$} &
\best{.852$_{\pm.016}$} &
\best{.869$_{\pm.017}$} &
\best{.771$_{\pm.023}$} &
\best{.452$_{\pm.014}$} &
\best{.477$_{\pm.020}$} &
\best{.312$_{\pm.020}$} &
\best{.445$_{\pm.016}$} &
\best{.490$_{\pm.010}$} &
\best{.447$_{\pm.017}$} &
\best{.331$_{\pm.009}$} &
\best{.383$_{\pm.014}$} &
\best{.417$_{\pm.007}$} &
\best{.833$_{\pm.012}$} &
\best{.492$_{\pm.015}$} &
\best{.755$_{\pm.009}$} \\
\bottomrule[1pt]
\end{tabular}}
\caption{Performance under standard settings on 4 MAG datasets. The \best{best}, \snd{second}, and \trd{third} results are marked in bold, underline, and italics, respectively.}
\label{tab:main}
\end{table*}

\begin{algorithm}[htbp]
\fontsize{8pt}{4pt}\selectfont
\DontPrintSemicolon
\SetAlgoLined
\caption{Overall Procedure of \textsc{RHEA}}
\label{alg:rhea}

\KwInput{Multimodal graph $G=(\mathcal V,\mathbf A,\{\mathbf X^{(m)}\})$, cluster number $K$, reliability threshold $\gamma$, temperatures $\tau$, $\tau_a$.}

initialize modality encoders $\{f_m\}$ and cluster prototypes $\{\boldsymbol{\mu}_k\}$;\\

\While{not converged}{

\tcc{Neighbor-Consensus Reliability Estimation}

compute modality embeddings $\{\mathbf Z^{(m)}\}$;\\

compute neighbor agreement $a_{i,m}$ and modality reliability $\rho_{i,m}$ via Eq.~(\ref{eq:agree})--(\ref{eq:rho});\\

\tcc{Reliability-Aware Representation Learning}

identify unreliable or missing modalities via Eq.~(\ref{eq:gate});\\

reconstruct unreliable representations and estimate reconstruction confidence via Eq.~(\ref{eq:recon});\\

obtain reliability-aware fused embeddings $\mathbf z$ via Eq.~(\ref{eq:fuse});\\

\tcc{Reliability-Aware Clustering}

compute topology-aware embeddings $\mathbf h$;\\

solve reliability-aware optimal transport via Eq.~(\ref{eq:ot});\\

compute neighbor-consensus assignments via Eq.~(\ref{eq:ncrc});\\

optimize the joint objective in Eq.~(\ref{eq:joint}) and update model parameters.
}

\KwOutput{Cluster assignments $\{\hat y_i\}_{i=1}^{N}$.}

\end{algorithm}

\section{Experiments}

In this section, we conduct a comprehensive evaluation of RHEA. We first describe the experimental settings (Sec.~\ref{sec: experimental setup}), and then answer the following research questions: \textbf{RQ1:} Is RHEA competitive with state-of-the-art baselines under clean and complete MAGs (Sec.~\ref{exp: rq1})? \textbf{RQ2:} Does RHEA consistently outperform existing methods when modalities are corrupted or partially missing (Sec.~\ref{exp: rq2})? \textbf{RQ3:} What are the individual contributions of each module in RHEA to the overall robustness gains (Sec.~\ref{exp: rq3})? \textbf{RQ4:} How stable are RHEA's performance gains under different hyperparameter settings and perturbation intensities (Sec.~\ref{exp: rq4})? \textbf{RQ5:} Does the reliability field track real corruption, and what does it reveal about how RHEA works (Sec.~\ref{exp: rq5})? Computational cost and complexity are analyzed in Appendix~\ref{app:complexity}.

\subsection{Experimental Setup}
\label{sec: experimental setup}

\paragraph{Datasets.} We evaluate RHEA on four public MAG datasets: the social network RedditS~\cite{RedditS}, two e-commerce networks (Toys and Grocery)~\cite{Amazon2018}, and the co-purchase graph Amazon~\cite{guo2025dmgc}. Detailed statistics and descriptions are in Appendix~\ref{app:datasets}.

\paragraph{Baselines.}
We compare RHEA against four categories of representative baselines. (1) \textit{Classical clustering methods}, which perform clustering solely based on node representations without exploiting graph topology, including KMeans. (2) \textit{General attributed graph clustering methods}, which leverage graph structure and node attributes but do not explicitly model multimodal reliability, including DFCN~\citep{tu2021dfcn}, DMoN~\citep{tsitsulin2023dmon}, MVGRL~\citep{hassani2020mvgrl} and S3GC~\citep{devvrit2022s3gc}. (3) \textit{Incomplete multi-view clustering methods}, which are specifically designed to handle missing-view scenarios, including COMPLETER~\citep{lin2021completer} and APADC~\citep{xu2022apadc}. (4) \textit{MAG clustering methods}, which jointly exploit graph topology and multiple modalities for clustering. These are our primary comparison methods, including DGF~\citep{zheng2025dgf} and DMGC~\citep{guo2025dmgc}. All baselines are implemented using their official codebases under an identical experimental protocol; per-baseline descriptions are in Appendix~\ref{app:baselines}. We report NMI, ACC, ARI, and F1 (defined in Appendix~\ref{app:metrics}) as the mean over five seeds under the protocol of Appendix~\ref{app:protocol}.

\subsection{Performance under Standard Setting}
\label{exp: rq1}

To answer \textbf{RQ1}, we evaluate clustering quality on all four benchmarks in the standard setting, where every
modality is complete and uncorrupted. The results are summarized in Table~\ref{tab:main}. We summarize our
observations as follows.

\paragraph{RHEA consistently achieves the best performance.} RHEA attains the best result on all four
datasets across every evaluation metric, demonstrating its effectiveness in the clean,
full-modality setting. On RedditS it clearly outperforms the strongest baseline DGF (NMI: $0.868$ vs.\ $0.836$;
ARI: $0.869$ vs.\ $0.811$), and on Grocery and Amazon it surpasses the second-best method by $0.034$ and $0.036$
in NMI, respectively. RHEA also maintains consistently low standard deviations, in contrast to methods such as
DMoN and APADC, which show high variance and on the harder datasets cluster little better than chance. This
stability indicates that the neighbor-consensus reliability estimation converges to a stable fusion solution (Appendix~\ref{app:ema}).

\begin{figure*}[t]
\centering
\includegraphics[width=\textwidth]{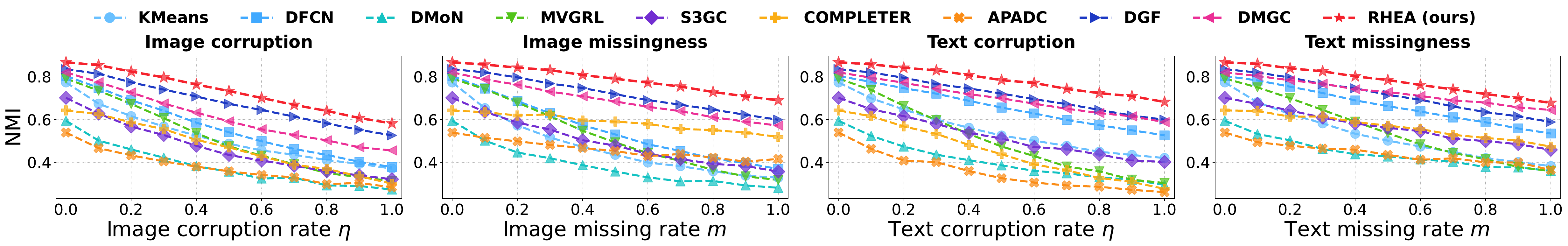}
\caption{NMI vs. perturbation severity on RedditS. Each panel varies one perturbation regime.}
\label{fig:degrade}
\end{figure*}

\paragraph{RHEA is particularly effective on complex MAGs.} The advantage of multimodal fusion is most
pronounced on benchmarks with many categories, such as Toys ($18$ clusters) and Grocery ($20$ clusters). On
Grocery, RHEA reaches an NMI of $0.490$, well above the deep attributed clusterer DFCN ($0.402$) and the
multimodal competitor DGF ($0.456$), showing its capacity to synthesize complementary information across views.
Amazon is a special case: with only three clusters and highly discriminative attributes, even strong baselines
such as KMeans ($0.358$) and DFCN ($0.381$) are competitive. Even there, RHEA still ranks first ($0.417$ NMI),
confirming that its reliability-guided fusion remains beneficial when simpler baselines are already strong.

\subsection{Robustness to Modality Perturbation}
\label{exp: rq2}

To evaluate \textbf{RQ2}, we degrade one modality while keeping the other intact. 
Table~\ref{tab:robust_nmi} reports NMI across five settings, including clean data, image and text corruption with $\eta=0.4$, and image and text missingness with $m=0.5$. 
Results for other metrics are in Appendix~\ref{app:permetric} (Tables~\ref{tab:robust_acc}--\ref{tab:robust_f1}).
Fig.~\ref{fig:degrade} illustrates performance trends as perturbation severity increases.

\paragraph{Overall performance.}
RHEA achieves the highest NMI across all configurations (4 datasets $\times$ 5 settings, 20 cases), consistently outperforming all baselines under both corruption and missingness. The average margin over DGF increases from $+0.04$ in clean conditions to $+0.07$ under image corruption, $+0.08$ under image missingness, and $\sim+0.07$ under text perturbations, indicating stronger robustness under modality degradation.

\paragraph{Robustness under increasing perturbation.}
All methods degrade as perturbation severity increases, while RHEA exhibits the slowest performance drop. On RedditS, its NMI decreases from $0.868$ to $0.729$ when $m=0.8$, whereas DGF drops from $0.836$ to $0.647$, increasing the performance gap from $+0.03$ to $+0.08$. Similar trends hold across datasets and perturbation types, as shown in Figure~\ref{fig:degrade}. These results indicate that modeling node-wise reliability improves stability under severe modality degradation.

\begin{table}[t]
\centering
\setlength{\tabcolsep}{4pt}
\renewcommand{\arraystretch}{1.05}
\resizebox{\columnwidth}{!}{
\begin{tabular}{ll c cc cc}
\toprule[1.0pt]
\multirow{2}{*}{\textbf{Dataset}} & \multirow{2}{*}{\textbf{Method}} & \multirow{2}{*}{\textbf{Standard}} & \multicolumn{2}{c}{\textbf{Image Pert.}} & \multicolumn{2}{c}{\textbf{Text Pert.}} \\
\cmidrule(lr){4-5}\cmidrule(lr){6-7}
 & & & \textbf{Corr.} & \textbf{Miss.} & \textbf{Corr.} & \textbf{Miss.} \\
\midrule[0.1pt]
\multirow{10}{*}{RedditS} & KMeans & .774$_{\pm.004}$ & .525$_{\pm.015}$ & .436$_{\pm.010}$ & .562$_{\pm.011}$ & .502$_{\pm.013}$ \\
 & DFCN & .804$_{\pm.003}$ & .586$_{\pm.011}$ & .531$_{\pm.010}$ & .688$_{\pm.013}$ & .663$_{\pm.010}$ \\
 & DMoN & .595$_{\pm.134}$ & .383$_{\pm.157}$ & .358$_{\pm.179}$ & .411$_{\pm.177}$ & .428$_{\pm.160}$ \\
 & MVGRL & .792$_{\pm.006}$ & .537$_{\pm.012}$ & .494$_{\pm.016}$ & .533$_{\pm.009}$ & .537$_{\pm.009}$ \\
 & S3GC & .702$_{\pm.012}$ & .478$_{\pm.021}$ & .479$_{\pm.024}$ & .538$_{\pm.026}$ & .562$_{\pm.027}$ \\
 & COMPLETER & .643$_{\pm.013}$ & .506$_{\pm.028}$ & .591$_{\pm.023}$ & .482$_{\pm.031}$ & .572$_{\pm.028}$ \\
 & APADC & .540$_{\pm.056}$ & .381$_{\pm.067}$ & .454$_{\pm.073}$ & .360$_{\pm.084}$ & .436$_{\pm.102}$ \\
 & DGF & \snd{.836$_{\pm.003}$} & \snd{.708$_{\pm.007}$} & \snd{.719$_{\pm.009}$} & \snd{.746$_{\pm.005}$} & \trd{.716$_{\pm.008}$} \\
 & DMGC & \trd{.821$_{\pm.005}$} & \trd{.633$_{\pm.010}$} & \trd{.685$_{\pm.009}$} & \trd{.719$_{\pm.013}$} & \snd{.729$_{\pm.015}$} \\
 & RHEA (ours) & \best{.868$_{\pm.008}$} & \best{.764$_{\pm.017}$} & \best{.791$_{\pm.015}$} & \best{.808$_{\pm.016}$} & \best{.785$_{\pm.013}$} \\
\midrule[0.1pt]
\multirow{10}{*}{Toys} & KMeans & .308$_{\pm.004}$ & .224$_{\pm.014}$ & .183$_{\pm.007}$ & .235$_{\pm.013}$ & .211$_{\pm.008}$ \\
 & DFCN & .430$_{\pm.008}$ & .302$_{\pm.013}$ & .269$_{\pm.019}$ & .354$_{\pm.016}$ & .340$_{\pm.014}$ \\
 & DMoN & .130$_{\pm.108}$ & .085$_{\pm.042}$ & .079$_{\pm.040}$ & .093$_{\pm.046}$ & .094$_{\pm.047}$ \\
 & MVGRL & .402$_{\pm.010}$ & .304$_{\pm.015}$ & .279$_{\pm.017}$ & .303$_{\pm.018}$ & .298$_{\pm.018}$ \\
 & S3GC & .310$_{\pm.014}$ & .213$_{\pm.027}$ & .213$_{\pm.021}$ & .236$_{\pm.029}$ & .256$_{\pm.023}$ \\
 & COMPLETER & .343$_{\pm.016}$ & .271$_{\pm.023}$ & .322$_{\pm.025}$ & .259$_{\pm.028}$ & .308$_{\pm.025}$ \\
 & APADC & .262$_{\pm.018}$ & .195$_{\pm.035}$ & .227$_{\pm.033}$ & .183$_{\pm.026}$ & .222$_{\pm.029}$ \\
 & DGF & \snd{.434$_{\pm.011}$} & \snd{.359$_{\pm.025}$} & \trd{.362$_{\pm.025}$} & \trd{.379$_{\pm.019}$} & \trd{.370$_{\pm.018}$} \\
 & DMGC & \trd{.432$_{\pm.010}$} & \trd{.342$_{\pm.016}$} & \snd{.365$_{\pm.017}$} & \snd{.382$_{\pm.017}$} & \snd{.389$_{\pm.017}$} \\
 & RHEA (ours) & \best{.452$_{\pm.014}$} & \best{.409$_{\pm.020}$} & \best{.416$_{\pm.021}$} & \best{.420$_{\pm.025}$} & \best{.411$_{\pm.022}$} \\
\midrule[0.1pt]
\multirow{10}{*}{Grocery} & KMeans & .202$_{\pm.006}$ & .130$_{\pm.014}$ & .107$_{\pm.014}$ & .143$_{\pm.009}$ & .125$_{\pm.014}$ \\
 & DFCN & .402$_{\pm.008}$ & .292$_{\pm.015}$ & .262$_{\pm.018}$ & .343$_{\pm.014}$ & .325$_{\pm.016}$ \\
 & DMoN & .124$_{\pm.098}$ & .078$_{\pm.039}$ & .072$_{\pm.036}$ & .082$_{\pm.041}$ & .084$_{\pm.042}$ \\
 & MVGRL & .386$_{\pm.009}$ & .266$_{\pm.018}$ & .251$_{\pm.016}$ & .267$_{\pm.017}$ & .263$_{\pm.024}$ \\
 & S3GC & .318$_{\pm.010}$ & .214$_{\pm.026}$ & .219$_{\pm.016}$ & .242$_{\pm.020}$ & .254$_{\pm.019}$ \\
 & COMPLETER & .370$_{\pm.033}$ & .283$_{\pm.063}$ & .337$_{\pm.060}$ & .268$_{\pm.041}$ & .321$_{\pm.048}$ \\
 & APADC & .283$_{\pm.028}$ & .215$_{\pm.058}$ & .249$_{\pm.042}$ & .201$_{\pm.041}$ & .243$_{\pm.041}$ \\
 & DGF & \snd{.456$_{\pm.003}$} & \snd{.372$_{\pm.013}$} & \trd{.375$_{\pm.008}$} & \trd{.387$_{\pm.006}$} & \trd{.378$_{\pm.009}$} \\
 & DMGC & \trd{.438$_{\pm.005}$} & \trd{.353$_{\pm.013}$} & \snd{.384$_{\pm.010}$} & \snd{.387$_{\pm.011}$} & \snd{.400$_{\pm.010}$} \\
 & RHEA (ours) & \best{.490$_{\pm.010}$} & \best{.454$_{\pm.022}$} & \best{.473$_{\pm.017}$} & \best{.474$_{\pm.024}$} & \best{.470$_{\pm.025}$} \\
\midrule[0.1pt]
\multirow{10}{*}{Amazon} & KMeans & \trd{.358$_{\pm.003}$} & .248$_{\pm.007}$ & .204$_{\pm.009}$ & .268$_{\pm.007}$ & .241$_{\pm.009}$ \\
 & DFCN & \snd{.381$_{\pm.009}$} & .260$_{\pm.013}$ & .232$_{\pm.019}$ & \trd{.306$_{\pm.019}$} & .284$_{\pm.016}$ \\
 & DMoN & .347$_{\pm.051}$ & .228$_{\pm.063}$ & .212$_{\pm.070}$ & .240$_{\pm.063}$ & .247$_{\pm.087}$ \\
 & MVGRL & .333$_{\pm.010}$ & .239$_{\pm.025}$ & .223$_{\pm.020}$ & .241$_{\pm.022}$ & .240$_{\pm.021}$ \\
 & S3GC & .301$_{\pm.012}$ & .203$_{\pm.029}$ & .201$_{\pm.023}$ & .226$_{\pm.022}$ & .239$_{\pm.025}$ \\
 & COMPLETER & .189$_{\pm.089}$ & .147$_{\pm.074}$ & .172$_{\pm.086}$ & .140$_{\pm.070}$ & .168$_{\pm.084}$ \\
 & APADC & .064$_{\pm.025}$ & .049$_{\pm.024}$ & .057$_{\pm.028}$ & .046$_{\pm.023}$ & .054$_{\pm.027}$ \\
 & DGF & .341$_{\pm.011}$ & \snd{.296$_{\pm.020}$} & \snd{.298$_{\pm.017}$} & \snd{.309$_{\pm.019}$} & \trd{.294$_{\pm.026}$} \\
 & DMGC & .352$_{\pm.008}$ & \trd{.274$_{\pm.016}$} & \trd{.295$_{\pm.016}$} & .304$_{\pm.020}$ & \snd{.316$_{\pm.021}$} \\
 & RHEA (ours) & \best{.417$_{\pm.007}$} & \best{.372$_{\pm.010}$} & \best{.387$_{\pm.018}$} & \best{.383$_{\pm.019}$} & \best{.376$_{\pm.017}$} \\
\bottomrule[1.0pt]
\end{tabular}}
\caption{Robustness (NMI) across standard, image-, and text-modality perturbation. Image/text corruption $\eta{=}0.4$, missing $m{=}0.5$.}
\label{tab:robust_nmi}
\end{table}

\subsection{Ablation Study}
\label{exp: rq3}

To answer \textbf{RQ3}, we remove each component of RHEA in turn and report mean NMI across the four datasets
under the clean, corruption, and missing-modality regimes (Table~\ref{tab:ablation}). Each component contributes
most in the regime it is designed for.

\begin{table}[t]
\centering
\setlength{\tabcolsep}{5pt}
\renewcommand{\arraystretch}{1.15}
\resizebox{\columnwidth}{!}{%
\begin{tabular}{l ccc}
\toprule[1.0pt]
\textbf{Variant} & \textbf{Standard} & \textbf{Corr.} & \textbf{Miss.} \\
\midrule[0.1pt]
RHEA (full) & \best{.557$_{\pm.011}$} & \best{.500$_{\pm.016}$} & \best{.517$_{\pm.015}$} \\ 
$-$ Reconstruction & \trd{.553$_{\pm.010}$} & .454$_{\pm.021}$ & .456$_{\pm.023}$ \\ 
$-$ Reliability Gating & .536$_{\pm.017}$ & .450$_{\pm.022}$ & .502$_{\pm.014}$ \\ 
$-$ Reliability Fusion & .550$_{\pm.013}$ & .484$_{\pm.015}$ & .505$_{\pm.019}$ \\  
$-$ Conf. OT Marginal & \snd{.554$_{\pm.009}$} & \snd{.490$_{\pm.012}$} & \snd{.510$_{\pm.013}$} \\  
$-$ NCRC & .538$_{\pm.020}$ & \trd{.487$_{\pm.017}$} & \trd{.508$_{\pm.016}$} \\ 
Backbone (all off) & .518$_{\pm.012}$ & .437$_{\pm.019}$ & .442$_{\pm.021}$ \\
\bottomrule[1.0pt]
\end{tabular}}
\caption{Component ablation across regimes (mean NMI $\pm$ std; four datasets, five seeds; image corruption $\eta{=}0.4$, missing $m{=}0.5$). Each row removes one component from full RHEA; the last removes all (only backbone).}
\label{tab:ablation}
\end{table}

\noindent\textbf{Reconstruction dominates the missing regime and the reliability gate the corrupted one.} Removing
the gated reconstruction lowers mean NMI by $0.061$ under missingness but by only $0.004$ when modalities are
clean, the sharpest regime contrast of any component, and recovers most of RHEA's advantage over the bare backbone
there, which is why its lead over DGF is widest under missing modalities (Sec.~\ref{exp: rq2}). The reliability
gate makes that reconstruction selective: reconstructing every modality rather than only the ones the field flags
as unreliable costs $0.050$ NMI under corruption, the largest gate-sensitive drop, because overwriting an
already-reliable modality injects noise; under missingness the gate matters less ($0.015$), since an absent
modality is always flagged. The field thus helps mainly by deciding which modality to repair, a role examined
further in Sec.~\ref{exp: rq5}.

\noindent\textbf{The clean base comes from assignment distillation over the shared backbone.} Neighbor-consensus
assignment distillation (NCRC) is the most useful clean-regime component outside the repair pathway ($0.019$ NMI
when removed); with every component off, NMI falls to the DGF-level backbone ($0.518$/$0.437$/$0.442$ for
clean/corruption/missing), so the clean-setting accuracy rests on NCRC over that backbone while the repair
mechanisms above add the missing-modality robustness on top. Reliability-weighted fusion and the confidence-shrunk
optimal-transport marginal only refine the result (each at most $0.016$); the choice of reconstruction operator
and the generality of the mechanism across text, joint, and asymmetric degradation are in
Appendix~\ref{app:more_exp}.
\subsection{Sensitivity Analysis}
\label{exp: rq4}

To answer \textbf{RQ4}, we check that the gains are stable across RHEA's four main hyperparameters: the entropic-OT
strength $\epsilon$, the number of Sinkhorn iterations, the reconstruction gate threshold $\gamma$, and the
reliability temperature $\tau$ (Table~\ref{tab:sensitivity}). We sweep each in the clean setting and under missing
modalities ($m{=}0.5$), the regime where RHEA's advantage is largest and its components contribute most
(Sec.~\ref{exp: rq2} and~\ref{exp: rq3}); the corruption regime is intermediate, so these two settings bracket the
operating range.
Across every sweep NMI stays within a narrow band around the default: a $20$-fold change in $\epsilon$ shifts it
by at most $0.020$, Sinkhorn iterations and $\tau$ by at most $0.006$. The gate $\gamma$ (swept over $[0.30,0.65]$
around its default $0.45$) is the most regime-sensitive knob---its band widens from $0.008$ (clean) to $0.022$
under missingness, where it decides which modality to repair---yet the default still sits near the top of the band
rather than at a sharp optimum, so RHEA needs no per-dataset tuning; homophily sensitivity is in
Appendix~\ref{app:more_sensitivity}.

\begin{table}[t]
\centering
\setlength{\tabcolsep}{5pt}
\renewcommand{\arraystretch}{1.15}
\resizebox{\columnwidth}{!}{%
\begin{tabular}{l c cc}
\toprule[1.0pt]
\textbf{Hyperparameter} & \textbf{Sweep} & \textbf{Standard} & \textbf{Miss.} \\
\midrule[0.1pt]

Entropic strength $\epsilon$ ($0.05$) & $20\times$ & $[.851,.871]$ & $[.775,.793]$ \\
Sinkhorn iterations ($20$) & $5\!-\!20$ & $[.864,.868]$ & $[.785,.791]$ \\
Gate threshold $\gamma$ ($0.45$) & $[.30,.65]$ & $[.861,.869]$ & $[.770,.792]$ \\
Reliability temp.\ $\tau$ ($1$) & $[.5,2]$ & $[.865,.869]$ & $[.786,.792]$ \\
\bottomrule[1.0pt]
\end{tabular}}
\caption{Hyperparameter sensitivity on RedditS (NMI). Each row sweeps one hyperparameter over the listed
range with all others fixed at their default (in parentheses); clean and missing $m{=}0.5$ regimes.}
\label{tab:sensitivity}
\end{table}

\subsection{Interpretability Investigation}
\label{exp: rq5}

\begin{figure}[t]
\centering

\includegraphics[width=\columnwidth]{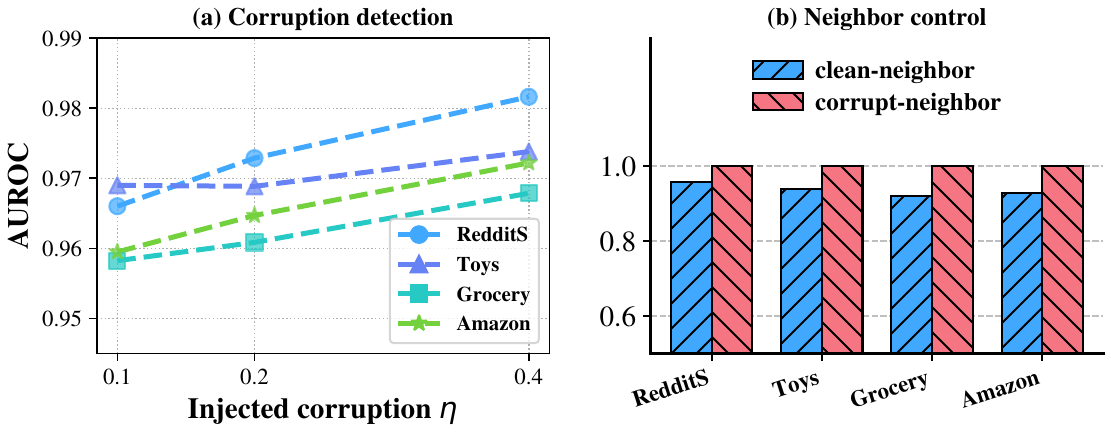}
\caption{Synthetic-corruption recoverability of the reliability field (corruption regime). (a)~AUROC of
$(1-\rho_{i,\mathrm{img}})$ vs.\ injected corruption $\eta$, per dataset. (b)~AUROC stratified by clean- vs.\
corrupt-neighbor nodes ($\eta{=}0.4$).}
\label{fig:ident}
\end{figure}

\begin{figure}[t]
\centering
\includegraphics[width=0.95\columnwidth]{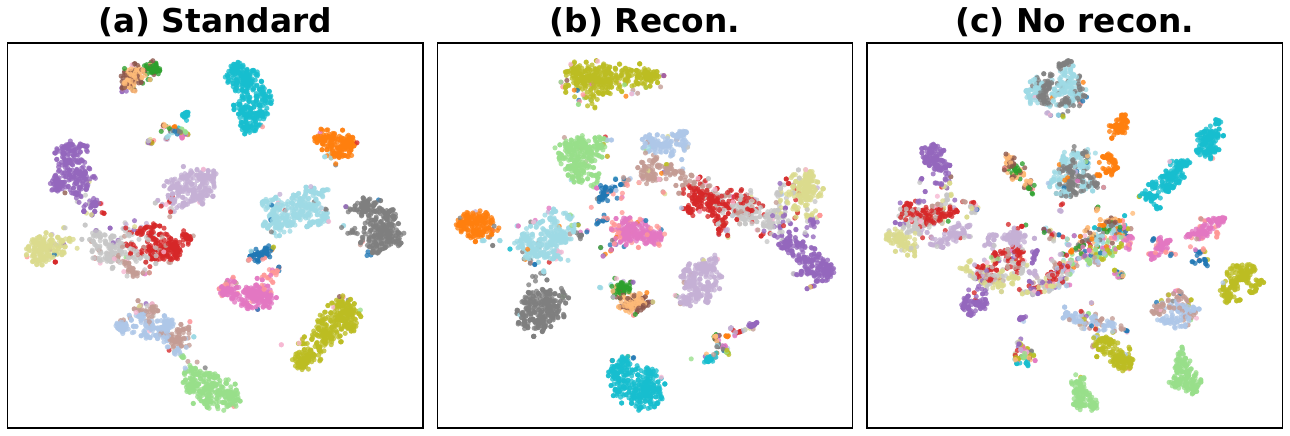}
\caption{t-SNE of RHEA fused embeddings on RedditS. (a)~standard setting; (b)~missing $m{=}0.5$ with
reconstruction; (c)~missing $m{=}0.5$ without reconstruction.}
\label{fig:tsne}
\end{figure}

To answer \textbf{RQ5}, we test what the reliability field learns and how it shapes the embedding through synthetic-corruption recoverability: we inject corruption into a known subset
of nodes and ask, without supervision, whether the field recovers it. The detector $(1-\rho_{i,\mathrm{img}})$
separates injected-corrupt from clean nodes at AUROC $0.958$--$0.982$ across all four benchmarks, rising with the
corruption severity $\eta$ (Fig.~\ref{fig:ident}a); restricting to the clean-neighbor stratum, where a majority
vote has nothing corrupt to imitate, it still scores $0.92$--$0.96$ (Fig.~\ref{fig:ident}b), so it reads each
node's own corruption rather than echoing its neighborhood. Accurate detection turns into a clustering gain only
through repair: the improvement under corruption is smaller and more dataset-dependent than under missingness
(Sec.~\ref{exp: rq2}), because repair acts only on gate-flagged nodes: an absent modality is always flagged and
replaced by the neighbor average (Eq.~\ref{eq:recon}), while sub-threshold corruption is retained. This is visible in the embedding
(Fig.~\ref{fig:tsne}): clusters are well separated in the standard setting, preserved under $50\%$ image
missingness when reconstruction is on, and collapse together when it is off, corroborating the ablation
(Sec.~\ref{exp: rq3}) that graph-neighbor reconstruction underlies the missing-modality robustness.

\section{Conclusion}

We presented RHEA, a multimodal graph clustering framework that estimates node-level modality reliability from neighborhood consensus, enabling adaptive feature reconstruction, reliability-aware reweighting of corrupted modalities, and topology-aware cluster assignment. Extensive experiments across diverse benchmarks and evaluation settings demonstrate that RHEA consistently outperforms competitive baselines, with particularly pronounced improvements under missing and corrupted modalities. These findings highlight the importance of modeling node-specific, graph-induced reliability for robust multimodal representation learning and clustering, suggesting that local structural context provides an effective signal for identifying unreliable observations. Future work will investigate extending the framework to low-homophily and heterophilous graphs.

\bibliography{references}

\newpage
\appendix

\section{Robustness to Modality Perturbation (ACC, ARI, F1)}
\label{app:permetric}

The NMI robustness results are reported in Table~\ref{tab:robust_nmi} (Sec.~\ref{exp: rq2}).
Tables~\ref{tab:robust_acc}--\ref{tab:robust_f1} further present the corresponding ACC, ARI, and F1 scores under the same evaluation protocol, including identical perturbation settings ($\eta = 0.4$, $m = 0.5$) and ranking marks. The \textbf{Standard} column reproduces the clean-performance results from the main comparison in Table~\ref{tab:main}. As observed, across all metrics and all settings, RHEA consistently achieves the best performance in every case, in agreement with the NMI results. This consistency suggests that the observed robustness gains are not metric-specific, but instead reflect a general and stable improvement under perturbations.

\begin{table}[t]
\centering
\setlength{\tabcolsep}{4pt}
\renewcommand{\arraystretch}{1.05}
\resizebox{\columnwidth}{!}{
\begin{tabular}{ll c cc cc}
\toprule[1.0pt]
\multirow{2}{*}{\textbf{Dataset}} & \multirow{2}{*}{\textbf{Method}} & \multirow{2}{*}{\textbf{Standard}} & \multicolumn{2}{c}{\textbf{Image Pert.}} & \multicolumn{2}{c}{\textbf{Text Pert.}} \\
\cmidrule(lr){4-5}\cmidrule(lr){6-7}
 & & & \textbf{Corr.} & \textbf{Miss.} & \textbf{Corr.} & \textbf{Miss.} \\
\midrule[0.1pt]
\multirow{10}{*}{RedditS} & KMeans & .737$_{\pm.010}$ & .498$_{\pm.019}$ & .419$_{\pm.025}$ & .534$_{\pm.018}$ & .486$_{\pm.024}$ \\
 & DFCN & \snd{.812$_{\pm.011}$} & .594$_{\pm.015}$ & .543$_{\pm.024}$ & .697$_{\pm.022}$ & .660$_{\pm.022}$ \\
 & DMoN & .531$_{\pm.182}$ & .332$_{\pm.166}$ & .319$_{\pm.160}$ & .366$_{\pm.183}$ & .370$_{\pm.185}$ \\
 & MVGRL & .781$_{\pm.012}$ & .530$_{\pm.021}$ & .497$_{\pm.028}$ & .524$_{\pm.027}$ & .516$_{\pm.019}$ \\
 & S3GC & .651$_{\pm.020}$ & .446$_{\pm.030}$ & .444$_{\pm.026}$ & .498$_{\pm.037}$ & .528$_{\pm.025}$ \\
 & COMPLETER & .534$_{\pm.026}$ & .422$_{\pm.038}$ & .496$_{\pm.038}$ & .402$_{\pm.039}$ & .483$_{\pm.056}$ \\
 & APADC & .339$_{\pm.086}$ & .238$_{\pm.119}$ & .286$_{\pm.143}$ & .231$_{\pm.115}$ & .276$_{\pm.138}$ \\
 & DGF & \trd{.807$_{\pm.014}$} & \snd{.676$_{\pm.027}$} & \snd{.689$_{\pm.030}$} & \snd{.712$_{\pm.030}$} & \trd{.678$_{\pm.023}$} \\
 & DMGC & .799$_{\pm.013}$ & \trd{.625$_{\pm.020}$} & \trd{.664$_{\pm.021}$} & \trd{.701$_{\pm.024}$} & \snd{.708$_{\pm.030}$} \\
 & RHEA (ours) & \best{.852$_{\pm.016}$} & \best{.762$_{\pm.028}$} & \best{.784$_{\pm.029}$} & \best{.781$_{\pm.036}$} & \best{.790$_{\pm.030}$} \\
\midrule[0.1pt]
\multirow{10}{*}{Toys} & KMeans & .353$_{\pm.004}$ & .250$_{\pm.008}$ & .206$_{\pm.013}$ & .271$_{\pm.014}$ & .242$_{\pm.014}$ \\
 & DFCN & \snd{.441$_{\pm.009}$} & .307$_{\pm.021}$ & .279$_{\pm.023}$ & \trd{.363$_{\pm.014}$} & .344$_{\pm.020}$ \\
 & DMoN & .177$_{\pm.067}$ & .115$_{\pm.058}$ & .109$_{\pm.054}$ & .128$_{\pm.064}$ & .127$_{\pm.063}$ \\
 & MVGRL & .415$_{\pm.012}$ & .316$_{\pm.022}$ & .289$_{\pm.023}$ & .314$_{\pm.029}$ & .308$_{\pm.018}$ \\
 & S3GC & .335$_{\pm.015}$ & .223$_{\pm.022}$ & .225$_{\pm.021}$ & .254$_{\pm.032}$ & .273$_{\pm.030}$ \\
 & COMPLETER & .340$_{\pm.013}$ & .272$_{\pm.023}$ & .311$_{\pm.025}$ & .258$_{\pm.026}$ & .307$_{\pm.023}$ \\
 & APADC & .246$_{\pm.019}$ & .179$_{\pm.025}$ & .213$_{\pm.030}$ & .174$_{\pm.035}$ & .203$_{\pm.037}$ \\
 & DGF & \trd{.424$_{\pm.018}$} & \snd{.344$_{\pm.034}$} & \trd{.354$_{\pm.031}$} & .361$_{\pm.032}$ & \trd{.360$_{\pm.035}$} \\
 & DMGC & .417$_{\pm.015}$ & \trd{.333$_{\pm.031}$} & \snd{.356$_{\pm.021}$} & \snd{.372$_{\pm.023}$} & \snd{.376$_{\pm.030}$} \\
 & RHEA (ours) & \best{.477$_{\pm.020}$} & \best{.434$_{\pm.038}$} & \best{.443$_{\pm.029}$} & \best{.446$_{\pm.036}$} & \best{.434$_{\pm.029}$} \\
\midrule[0.1pt]
\multirow{10}{*}{Grocery} & KMeans & .248$_{\pm.018}$ & .158$_{\pm.038}$ & .134$_{\pm.024}$ & .174$_{\pm.028}$ & .156$_{\pm.035}$ \\
 & DFCN & .406$_{\pm.013}$ & .291$_{\pm.021}$ & .265$_{\pm.025}$ & .338$_{\pm.028}$ & .330$_{\pm.026}$ \\
 & DMoN & .193$_{\pm.043}$ & .120$_{\pm.060}$ & .112$_{\pm.056}$ & .131$_{\pm.061}$ & .135$_{\pm.068}$ \\
 & MVGRL & .392$_{\pm.011}$ & .271$_{\pm.016}$ & .253$_{\pm.015}$ & .267$_{\pm.017}$ & .269$_{\pm.021}$ \\
 & S3GC & .341$_{\pm.012}$ & .234$_{\pm.020}$ & .228$_{\pm.023}$ & .262$_{\pm.024}$ & .274$_{\pm.025}$ \\
 & COMPLETER & .381$_{\pm.025}$ & .295$_{\pm.046}$ & \trd{.347$_{\pm.036}$} & .287$_{\pm.033}$ & \trd{.343$_{\pm.034}$} \\
 & APADC & .285$_{\pm.033}$ & .209$_{\pm.049}$ & .248$_{\pm.062}$ & .206$_{\pm.040}$ & .239$_{\pm.067}$ \\
 & DGF & \snd{.411$_{\pm.005}$} & \snd{.332$_{\pm.017}$} & .335$_{\pm.012}$ & \trd{.351$_{\pm.012}$} & .336$_{\pm.009}$ \\
 & DMGC & \trd{.408$_{\pm.006}$} & \trd{.331$_{\pm.013}$} & \snd{.351$_{\pm.017}$} & \snd{.366$_{\pm.014}$} & \snd{.375$_{\pm.012}$} \\
 & RHEA (ours) & \best{.447$_{\pm.017}$} & \best{.409$_{\pm.024}$} & \best{.423$_{\pm.031}$} & \best{.433$_{\pm.033}$} & \best{.419$_{\pm.034}$} \\
\midrule[0.1pt]
\multirow{10}{*}{Amazon} & KMeans & \snd{.799$_{\pm.001}$} & .534$_{\pm.005}$ & .444$_{\pm.007}$ & .590$_{\pm.005}$ & .515$_{\pm.009}$ \\
 & DFCN & \trd{.763$_{\pm.010}$} & .512$_{\pm.021}$ & .462$_{\pm.025}$ & .601$_{\pm.017}$ & .581$_{\pm.014}$ \\
 & DMoN & .710$_{\pm.096}$ & .456$_{\pm.122}$ & .435$_{\pm.184}$ & .495$_{\pm.186}$ & .509$_{\pm.118}$ \\
 & MVGRL & .725$_{\pm.012}$ & .517$_{\pm.023}$ & .489$_{\pm.024}$ & .522$_{\pm.021}$ & .523$_{\pm.020}$ \\
 & S3GC & .701$_{\pm.015}$ & .467$_{\pm.018}$ & .462$_{\pm.030}$ & .513$_{\pm.026}$ & .546$_{\pm.033}$ \\
 & COMPLETER & .569$_{\pm.156}$ & .442$_{\pm.220}$ & .524$_{\pm.180}$ & .423$_{\pm.211}$ & .504$_{\pm.220}$ \\
 & APADC & .649$_{\pm.013}$ & .495$_{\pm.023}$ & .585$_{\pm.031}$ & .466$_{\pm.026}$ & .564$_{\pm.021}$ \\
 & DGF & .740$_{\pm.013}$ & \snd{.629$_{\pm.018}$} & \snd{.651$_{\pm.027}$} & \snd{.670$_{\pm.027}$} & \trd{.643$_{\pm.029}$} \\
 & DMGC & .749$_{\pm.010}$ & \trd{.590$_{\pm.015}$} & \trd{.639$_{\pm.014}$} & \trd{.658$_{\pm.018}$} & \snd{.661$_{\pm.020}$} \\
 & RHEA (ours) & \best{.833$_{\pm.012}$} & \best{.738$_{\pm.027}$} & \best{.753$_{\pm.026}$} & \best{.764$_{\pm.024}$} & \best{.766$_{\pm.024}$} \\
\bottomrule[1.0pt]
\end{tabular}}
\caption{Robustness (ACC) across standard, image-, and text-modality perturbation. Image/text corruption $\eta{=}0.4$, missing $m{=}0.5$.}
\label{tab:robust_acc}
\end{table}

\begin{table}[t]
\centering
\setlength{\tabcolsep}{4pt}
\renewcommand{\arraystretch}{1.05}
\resizebox{\columnwidth}{!}{
\begin{tabular}{ll c cc cc}
\toprule[1.0pt]
\multirow{2}{*}{\textbf{Dataset}} & \multirow{2}{*}{\textbf{Method}} & \multirow{2}{*}{\textbf{Standard}} & \multicolumn{2}{c}{\textbf{Image Pert.}} & \multicolumn{2}{c}{\textbf{Text Pert.}} \\
\cmidrule(lr){4-5}\cmidrule(lr){6-7}
 & & & \textbf{Corr.} & \textbf{Miss.} & \textbf{Corr.} & \textbf{Miss.} \\
\midrule[0.1pt]
\multirow{10}{*}{RedditS} & KMeans & .689$_{\pm.017}$ & .469$_{\pm.034}$ & .392$_{\pm.031}$ & .506$_{\pm.036}$ & .454$_{\pm.031}$ \\
 & DFCN & .765$_{\pm.013}$ & .555$_{\pm.018}$ & .504$_{\pm.018}$ & .653$_{\pm.028}$ & .624$_{\pm.021}$ \\
 & DMoN & .451$_{\pm.185}$ & .289$_{\pm.144}$ & .276$_{\pm.138}$ & .313$_{\pm.156}$ & .326$_{\pm.163}$ \\
 & MVGRL & .742$_{\pm.015}$ & .495$_{\pm.025}$ & .461$_{\pm.024}$ & .491$_{\pm.028}$ & .490$_{\pm.023}$ \\
 & S3GC & .602$_{\pm.018}$ & .419$_{\pm.037}$ & .412$_{\pm.031}$ & .472$_{\pm.030}$ & .495$_{\pm.037}$ \\
 & COMPLETER & .476$_{\pm.033}$ & .388$_{\pm.050}$ & .453$_{\pm.052}$ & .364$_{\pm.062}$ & .442$_{\pm.050}$ \\
 & APADC & .239$_{\pm.059}$ & .169$_{\pm.085}$ & .198$_{\pm.099}$ & .161$_{\pm.080}$ & .193$_{\pm.076}$ \\
 & DGF & \snd{.811$_{\pm.013}$} & \snd{.686$_{\pm.025}$} & \snd{.711$_{\pm.022}$} & \snd{.723$_{\pm.019}$} & \snd{.697$_{\pm.020}$} \\
 & DMGC & \trd{.789$_{\pm.011}$} & \trd{.612$_{\pm.019}$} & \trd{.649$_{\pm.019}$} & \trd{.677$_{\pm.017}$} & \trd{.690$_{\pm.021}$} \\
 & RHEA (ours) & \best{.869$_{\pm.017}$} & \best{.766$_{\pm.028}$} & \best{.788$_{\pm.039}$} & \best{.815$_{\pm.039}$} & \best{.793$_{\pm.031}$} \\
\midrule[0.1pt]
\multirow{10}{*}{Toys} & KMeans & .171$_{\pm.001}$ & .122$_{\pm.008}$ & .100$_{\pm.005}$ & .131$_{\pm.006}$ & .116$_{\pm.006}$ \\
 & DFCN & .255$_{\pm.009}$ & .180$_{\pm.018}$ & .161$_{\pm.012}$ & .208$_{\pm.016}$ & .200$_{\pm.021}$ \\
 & DMoN & .058$_{\pm.057}$ & .038$_{\pm.019}$ & .035$_{\pm.018}$ & .041$_{\pm.020}$ & .041$_{\pm.021}$ \\
 & MVGRL & .236$_{\pm.011}$ & .174$_{\pm.017}$ & .163$_{\pm.022}$ & .176$_{\pm.017}$ & .174$_{\pm.021}$ \\
 & S3GC & .189$_{\pm.013}$ & .129$_{\pm.026}$ & .131$_{\pm.023}$ & .144$_{\pm.030}$ & .155$_{\pm.022}$ \\
 & COMPLETER & .188$_{\pm.014}$ & .150$_{\pm.027}$ & .175$_{\pm.026}$ & .142$_{\pm.030}$ & .167$_{\pm.028}$ \\
 & APADC & .071$_{\pm.016}$ & .053$_{\pm.025}$ & .061$_{\pm.031}$ & .050$_{\pm.025}$ & .060$_{\pm.028}$ \\
 & DGF & \snd{.268$_{\pm.011}$} & \snd{.222$_{\pm.015}$} & \trd{.222$_{\pm.016}$} & \trd{.229$_{\pm.025}$} & \trd{.222$_{\pm.024}$} \\
 & DMGC & \trd{.262$_{\pm.012}$} & \trd{.212$_{\pm.025}$} & \snd{.224$_{\pm.026}$} & \snd{.236$_{\pm.019}$} & \snd{.239$_{\pm.016}$} \\
 & RHEA (ours) & \best{.312$_{\pm.020}$} & \best{.283$_{\pm.036}$} & \best{.289$_{\pm.042}$} & \best{.288$_{\pm.032}$} & \best{.284$_{\pm.030}$} \\
\midrule[0.1pt]
\multirow{10}{*}{Grocery} & KMeans & .099$_{\pm.010}$ & .063$_{\pm.024}$ & .052$_{\pm.018}$ & .069$_{\pm.022}$ & .061$_{\pm.021}$ \\
 & DFCN & .255$_{\pm.011}$ & .184$_{\pm.022}$ & .163$_{\pm.014}$ & .212$_{\pm.020}$ & .206$_{\pm.025}$ \\
 & DMoN & .052$_{\pm.054}$ & .033$_{\pm.017}$ & .031$_{\pm.015}$ & .036$_{\pm.018}$ & .036$_{\pm.018}$ \\
 & MVGRL & .241$_{\pm.010}$ & .165$_{\pm.018}$ & .155$_{\pm.021}$ & .168$_{\pm.022}$ & .165$_{\pm.024}$ \\
 & S3GC & .176$_{\pm.011}$ & .118$_{\pm.024}$ & .120$_{\pm.014}$ & .135$_{\pm.020}$ & .141$_{\pm.018}$ \\
 & COMPLETER & .218$_{\pm.036}$ & .174$_{\pm.045}$ & .201$_{\pm.059}$ & .164$_{\pm.074}$ & .194$_{\pm.045}$ \\
 & APADC & .126$_{\pm.026}$ & .092$_{\pm.046}$ & .110$_{\pm.052}$ & .089$_{\pm.044}$ & .105$_{\pm.034}$ \\
 & DGF & \snd{.295$_{\pm.003}$} & \snd{.245$_{\pm.010}$} & \snd{.248$_{\pm.009}$} & \snd{.255$_{\pm.008}$} & \trd{.246$_{\pm.011}$} \\
 & DMGC & \trd{.283$_{\pm.004}$} & \trd{.229$_{\pm.013}$} & \trd{.246$_{\pm.014}$} & \trd{.252$_{\pm.013}$} & \snd{.261$_{\pm.009}$} \\
 & RHEA (ours) & \best{.331$_{\pm.009}$} & \best{.309$_{\pm.017}$} & \best{.319$_{\pm.016}$} & \best{.319$_{\pm.015}$} & \best{.311$_{\pm.017}$} \\
\midrule[0.1pt]
\multirow{10}{*}{Amazon} & KMeans & \snd{.465$_{\pm.003}$} & \trd{.314$_{\pm.012}$} & .262$_{\pm.010}$ & \snd{.344$_{\pm.007}$} & .305$_{\pm.007}$ \\
 & DFCN & \trd{.427$_{\pm.019}$} & .287$_{\pm.028}$ & .257$_{\pm.027}$ & \trd{.335$_{\pm.036}$} & \trd{.324$_{\pm.028}$} \\
 & DMoN & .363$_{\pm.112}$ & .233$_{\pm.116}$ & .221$_{\pm.111}$ & .256$_{\pm.128}$ & .259$_{\pm.129}$ \\
 & MVGRL & .401$_{\pm.013}$ & .286$_{\pm.016}$ & .274$_{\pm.025}$ & .293$_{\pm.023}$ & .286$_{\pm.022}$ \\
 & S3GC & .328$_{\pm.014}$ & .216$_{\pm.031}$ & .219$_{\pm.023}$ & .246$_{\pm.027}$ & .261$_{\pm.024}$ \\
 & COMPLETER & .109$_{\pm.201}$ & .086$_{\pm.043}$ & .100$_{\pm.050}$ & .080$_{\pm.040}$ & .096$_{\pm.048}$ \\
 & APADC & .045$_{\pm.023}$ & .034$_{\pm.017}$ & .040$_{\pm.020}$ & .032$_{\pm.016}$ & .039$_{\pm.019}$ \\
 & DGF & .372$_{\pm.020}$ & \snd{.316$_{\pm.041}$} & \trd{.315$_{\pm.035}$} & .325$_{\pm.033}$ & .316$_{\pm.035}$ \\
 & DMGC & .389$_{\pm.009}$ & .300$_{\pm.013}$ & \snd{.323$_{\pm.011}$} & .333$_{\pm.021}$ & \snd{.347$_{\pm.017}$} \\
 & RHEA (ours) & \best{.492$_{\pm.015}$} & \best{.431$_{\pm.028}$} & \best{.444$_{\pm.024}$} & \best{.460$_{\pm.031}$} & \best{.455$_{\pm.033}$} \\
\bottomrule[1.0pt]
\end{tabular}}
\caption{Robustness (ARI) across standard, image-, and text-modality perturbation. Image/text corruption $\eta{=}0.4$, missing $m{=}0.5$.}
\label{tab:robust_ari}
\end{table}

\begin{table}[t]
\centering
\setlength{\tabcolsep}{4pt}
\renewcommand{\arraystretch}{1.05}
\resizebox{\columnwidth}{!}{
\begin{tabular}{ll c cc cc}
\toprule[1.0pt]
\multirow{2}{*}{\textbf{Dataset}} & \multirow{2}{*}{\textbf{Method}} & \multirow{2}{*}{\textbf{Standard}} & \multicolumn{2}{c}{\textbf{Image Pert.}} & \multicolumn{2}{c}{\textbf{Text Pert.}} \\
\cmidrule(lr){4-5}\cmidrule(lr){6-7}
 & & & \textbf{Corr.} & \textbf{Miss.} & \textbf{Corr.} & \textbf{Miss.} \\
\midrule[0.1pt]
\multirow{10}{*}{RedditS} & KMeans & .651$_{\pm.006}$ & .434$_{\pm.014}$ & .359$_{\pm.018}$ & .472$_{\pm.013}$ & .425$_{\pm.014}$ \\
 & DFCN & \snd{.725$_{\pm.006}$} & .532$_{\pm.010}$ & .465$_{\pm.011}$ & \snd{.626$_{\pm.018}$} & .586$_{\pm.013}$ \\
 & DMoN & .394$_{\pm.214}$ & .252$_{\pm.126}$ & .236$_{\pm.118}$ & .271$_{\pm.135}$ & .279$_{\pm.139}$ \\
 & MVGRL & \trd{.701$_{\pm.009}$} & .472$_{\pm.012}$ & .445$_{\pm.015}$ & .470$_{\pm.019}$ & .471$_{\pm.019}$ \\
 & S3GC & .541$_{\pm.022}$ & .378$_{\pm.036}$ & .373$_{\pm.047}$ & .413$_{\pm.049}$ & .446$_{\pm.034}$ \\
 & COMPLETER & .400$_{\pm.030}$ & .317$_{\pm.049}$ & .370$_{\pm.050}$ & .298$_{\pm.054}$ & .354$_{\pm.048}$ \\
 & APADC & .227$_{\pm.073}$ & .165$_{\pm.083}$ & .191$_{\pm.096}$ & .156$_{\pm.078}$ & .183$_{\pm.091}$ \\
 & DGF & .695$_{\pm.031}$ & \snd{.590$_{\pm.046}$} & \snd{.591$_{\pm.052}$} & \trd{.604$_{\pm.061}$} & \trd{.586$_{\pm.054}$} \\
 & DMGC & .681$_{\pm.021}$ & \trd{.532$_{\pm.040}$} & \trd{.573$_{\pm.040}$} & .598$_{\pm.033}$ & \snd{.617$_{\pm.038}$} \\
 & RHEA (ours) & \best{.771$_{\pm.023}$} & \best{.685$_{\pm.035}$} & \best{.707$_{\pm.047}$} & \best{.718$_{\pm.037}$} & \best{.699$_{\pm.046}$} \\
\midrule[0.1pt]
\multirow{10}{*}{Toys} & KMeans & .315$_{\pm.003}$ & .225$_{\pm.012}$ & .186$_{\pm.010}$ & .243$_{\pm.007}$ & .221$_{\pm.011}$ \\
 & DFCN & \snd{.400$_{\pm.014}$} & .279$_{\pm.032}$ & .252$_{\pm.031}$ & \trd{.323$_{\pm.018}$} & .313$_{\pm.022}$ \\
 & DMoN & .103$_{\pm.097}$ & .067$_{\pm.034}$ & .062$_{\pm.031}$ & .073$_{\pm.037}$ & .074$_{\pm.037}$ \\
 & MVGRL & \trd{.382$_{\pm.010}$} & \trd{.285$_{\pm.017}$} & .266$_{\pm.018}$ & .284$_{\pm.019}$ & .286$_{\pm.020}$ \\
 & S3GC & .282$_{\pm.012}$ & .194$_{\pm.019}$ & .194$_{\pm.023}$ & .216$_{\pm.025}$ & .229$_{\pm.028}$ \\
 & COMPLETER & .277$_{\pm.011}$ & .215$_{\pm.023}$ & .252$_{\pm.022}$ & .206$_{\pm.022}$ & .247$_{\pm.027}$ \\
 & APADC & .179$_{\pm.023}$ & .129$_{\pm.033}$ & .153$_{\pm.042}$ & .126$_{\pm.033}$ & .147$_{\pm.030}$ \\
 & DGF & .372$_{\pm.015}$ & \snd{.311$_{\pm.022}$} & \snd{.315$_{\pm.030}$} & \snd{.330$_{\pm.030}$} & \trd{.316$_{\pm.023}$} \\
 & DMGC & .364$_{\pm.013}$ & .285$_{\pm.027}$ & \trd{.314$_{\pm.032}$} & .321$_{\pm.021}$ & \snd{.333$_{\pm.023}$} \\
 & RHEA (ours) & \best{.445$_{\pm.016}$} & \best{.394$_{\pm.033}$} & \best{.413$_{\pm.027}$} & \best{.421$_{\pm.023}$} & \best{.414$_{\pm.024}$} \\
\midrule[0.1pt]
\multirow{10}{*}{Grocery} & KMeans & .219$_{\pm.013}$ & .139$_{\pm.025}$ & .115$_{\pm.029}$ & .151$_{\pm.027}$ & .133$_{\pm.022}$ \\
 & DFCN & .328$_{\pm.012}$ & .236$_{\pm.025}$ & .213$_{\pm.021}$ & .273$_{\pm.021}$ & .265$_{\pm.019}$ \\
 & DMoN & .106$_{\pm.081}$ & .067$_{\pm.033}$ & .063$_{\pm.031}$ & .072$_{\pm.036}$ & .074$_{\pm.037}$ \\
 & MVGRL & .318$_{\pm.012}$ & .222$_{\pm.020}$ & .203$_{\pm.023}$ & .218$_{\pm.023}$ & .218$_{\pm.020}$ \\
 & S3GC & .264$_{\pm.013}$ & .181$_{\pm.020}$ & .180$_{\pm.020}$ & .201$_{\pm.025}$ & .216$_{\pm.026}$ \\
 & COMPLETER & .297$_{\pm.022}$ & .232$_{\pm.046}$ & .265$_{\pm.038}$ & .220$_{\pm.047}$ & .263$_{\pm.041}$ \\
 & APADC & .164$_{\pm.022}$ & .124$_{\pm.037}$ & .144$_{\pm.038}$ & .119$_{\pm.041}$ & .139$_{\pm.031}$ \\
 & DGF & \snd{.348$_{\pm.008}$} & \snd{.286$_{\pm.021}$} & \trd{.292$_{\pm.016}$} & \trd{.301$_{\pm.016}$} & \trd{.291$_{\pm.011}$} \\
 & DMGC & \trd{.340$_{\pm.008}$} & \trd{.277$_{\pm.016}$} & \snd{.298$_{\pm.021}$} & \snd{.307$_{\pm.017}$} & \snd{.321$_{\pm.018}$} \\
 & RHEA (ours) & \best{.383$_{\pm.014}$} & \best{.350$_{\pm.020}$} & \best{.371$_{\pm.030}$} & \best{.370$_{\pm.022}$} & \best{.365$_{\pm.022}$} \\
\midrule[0.1pt]
\multirow{10}{*}{Amazon} & KMeans & \trd{.716$_{\pm.002}$} & .486$_{\pm.010}$ & .404$_{\pm.007}$ & .523$_{\pm.009}$ & .471$_{\pm.010}$ \\
 & DFCN & \snd{.718$_{\pm.010}$} & .489$_{\pm.023}$ & .434$_{\pm.024}$ & .575$_{\pm.019}$ & .550$_{\pm.016}$ \\
 & DMoN & .645$_{\pm.089}$ & .415$_{\pm.131}$ & .391$_{\pm.158}$ & .461$_{\pm.132}$ & .463$_{\pm.172}$ \\
 & MVGRL & .690$_{\pm.010}$ & .497$_{\pm.020}$ & .475$_{\pm.014}$ & .503$_{\pm.023}$ & .498$_{\pm.023}$ \\
 & S3GC & .612$_{\pm.013}$ & .410$_{\pm.019}$ & .404$_{\pm.023}$ & .452$_{\pm.029}$ & .487$_{\pm.026}$ \\
 & COMPLETER & .462$_{\pm.148}$ & .366$_{\pm.183}$ & .431$_{\pm.215}$ & .350$_{\pm.175}$ & .410$_{\pm.183}$ \\
 & APADC & .330$_{\pm.034}$ & .246$_{\pm.059}$ & .297$_{\pm.050}$ & .235$_{\pm.049}$ & .284$_{\pm.047}$ \\
 & DGF & .673$_{\pm.011}$ & \snd{.571$_{\pm.018}$} & \snd{.586$_{\pm.021}$} & \snd{.602$_{\pm.016}$} & \trd{.586$_{\pm.019}$} \\
 & DMGC & .686$_{\pm.009}$ & \trd{.528$_{\pm.014}$} & \trd{.580$_{\pm.022}$} & \trd{.591$_{\pm.019}$} & \snd{.621$_{\pm.018}$} \\
 & RHEA (ours) & \best{.755$_{\pm.009}$} & \best{.669$_{\pm.023}$} & \best{.695$_{\pm.015}$} & \best{.705$_{\pm.020}$} & \best{.686$_{\pm.019}$} \\
\bottomrule[1.0pt]

\end{tabular}}
\caption{Robustness (F1) across standard, image-, and text-modality perturbation. Image/text corruption $\eta{=}0.4$, missing $m{=}0.5$.}
\label{tab:robust_f1}
\end{table}

\section{More Ablation Study}
\label{app:more_exp}

Beyond the component ablation study in Sec.~\ref{exp: rq3} (Table~\ref{tab:ablation}), we further conduct three analyses:
the choice of reconstruction operator, the generality of the reliability mechanism across modality corruption settings,
and a comparison against training-free graph imputation baselines.

\paragraph{Reconstruction operator.}
\label{app:operators}
The reconstruction module in Sec.~\ref{sec:recon} aggregates one-hop neighbors of each node from the reliable modality.
We evaluate alternative topology-aware denoising operators, including personalized PageRank (\textsc{PPR}, teleport $\alpha = 0.15$),
heat kernel diffusion (\textsc{Heat}), a similarity-gated one-hop operator that down-weights anti-correlated edges
(\textsc{Simgate}), and their compositions.

\begin{figure}[t]
\centering
\includegraphics[width=\columnwidth]{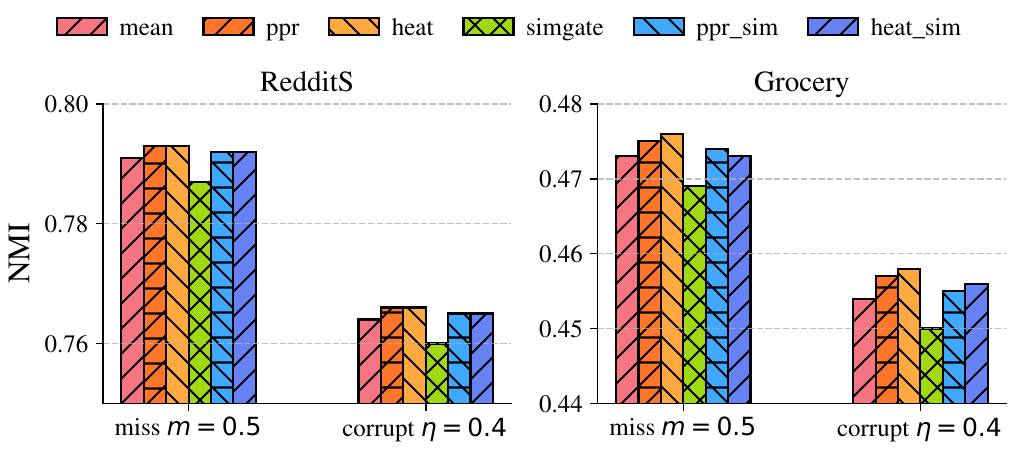}
\caption{Reconstruction operator ablation. NMI under missing ($m=0.5$) and corruption ($\eta=0.4$) settings for different reconstruction operators.}
\label{fig:operator}
\end{figure}

As shown in Fig.~\ref{fig:operator}, evaluated on RedditS and Grocery under missing ($m=0.5$) and corruption ($\eta=0.4$) settings
(averaged over five seeds), multi-hop diffusion methods (\textsc{PPR}/\textsc{Heat}) yield only marginal improvements over the
one-hop mean aggregator ($+0.002$ NMI on average). The similarity-gated variant does not provide consistent gains
($-0.004$ NMI on average). This is likely because similarity gating effectively filters low-confidence neighbors, while RHEA
already leverages per-node reliability via $c_i$ to identify informative donors. Since all alternatives lie within one standard
deviation of the mean and do not affect overall conclusions, we retain the one-hop mean as the default reconstruction operator.

\paragraph{Generality across modality regimes.}
\label{app:modality}
Since both the gating and reconstruction operate on the modality identified as unreliable, the mechanism is not restricted to a
specific modality. Table~\ref{tab:modality} evaluates RHEA against the backbone without reliability modeling (\textsc{Sinkhorn})
under different corruption regimes, including text-only corruption, joint corruption, and asymmetric modality degradation.

RHEA consistently improves over the baseline across all settings, with the largest gains under asymmetric corruption
($+0.21$ NMI on RedditS and $+0.06$ on Amazon) and joint corruption ($+0.10$ and $+0.06$). Even under single-modality text
perturbation, RHEA remains competitive ($+0.01$ on both datasets). These results indicate that reliability-aware reconstruction
is most beneficial when modality quality is uneven across nodes, consistent with the design assumption.

\begin{table}[t]
\centering
\setlength{\tabcolsep}{4pt}
{\small
\begin{tabular}{ll cc}
\toprule[1.0pt]
\textbf{Dataset} & \textbf{Regime} & \textbf{RHEA} & \textbf{\textsc{sinkhorn}} \\
\midrule[0.1pt]
\multirow{4}{*}{RedditS} & corrupt-text $\eta{=}.4$ & \best{.808$_{\pm.008}$} & .795$_{\pm.011}$ \\ 
 & corrupt-both $\eta{=}.4$ & \best{.700$_{\pm.010}$} & .600$_{\pm.022}$ \\ 
 & text-missing $m{=}.5$ & \best{.785$_{\pm.009}$} & .772$_{\pm.012}$ \\
 & asymmetric $m{=}.5$ & \best{.770$_{\pm.009}$} & .560$_{\pm.017}$ \\
\midrule[0.1pt]
\multirow{4}{*}{Amazon} & corrupt-text $\eta{=}.4$ & \best{.383$_{\pm.007}$} & .370$_{\pm.015}$ \\ 
 & corrupt-both $\eta{=}.4$ & \best{.330$_{\pm.009}$} & .270$_{\pm.020}$ \\ 
 & text-missing $m{=}.5$ & \best{.376$_{\pm.006}$} & .363$_{\pm.014}$ \\ 
 & asymmetric $m{=}.5$ & \best{.360$_{\pm.008}$} & .300$_{\pm.019}$ \\
\bottomrule[1.0pt]
\end{tabular}}
\caption{Generality across modality degradation regimes.}
\label{tab:modality}
\end{table}

\paragraph{Learned reconstruction vs.\ training-free graph imputation.}
Table~\ref{tab:impute} compares RHEA with training-free graph imputation strategies applied as preprocessing, including one-hop
mean, personalized PageRank, and heat kernel diffusion, while keeping the downstream backbone fixed.

RHEA consistently outperforms the best static imputation baseline (one-hop mean) by $+0.08$--$+0.09$ NMI. In contrast, multi-hop
diffusion methods do not improve over simple neighborhood averaging when used as fixed preprocessing, highlighting the benefit of
learning reconstruction jointly with reliability estimation.

\begin{table}[t]
\centering
\setlength{\tabcolsep}{4pt}
\renewcommand{\arraystretch}{1.0}
\resizebox{\columnwidth}{!}{%
\begin{tabular}{ll cccc}
\toprule[1.0pt]
\textbf{Dataset} & $m$ & \textbf{RHEA} & \textbf{impute-\textsc{mean}} & \textbf{impute-\textsc{ppr}} & \textbf{impute-\textsc{heat}} \\
\midrule[0.1pt]
\multirow{2}{*}{RedditS} & .3 & \best{.815$_{\pm.010}$} & \snd{.730$_{\pm.010}$} & .672$_{\pm.010}$ & \trd{.683$_{\pm.008}$} \\
 & .5 & \best{.791$_{\pm.013}$} & .700$_{\pm.009}$ & .601$_{\pm.008}$ & .631$_{\pm.012}$ \\
\midrule[0.1pt]
\multirow{2}{*}{Amazon} & .3 & \best{.400$_{\pm.006}$} & .322$_{\pm.007}$ & .310$_{\pm.005}$ & .318$_{\pm.007}$ \\
 & .5 & \best{.387$_{\pm.005}$} & .300$_{\pm.018}$ & .291$_{\pm.021}$ & .300$_{\pm.022}$ \\
\bottomrule[1.0pt]
\end{tabular}}
\caption{Comparison with training-free graph imputation under image-missing settings.}
\label{tab:impute}
\end{table}

\section{More Sensitivity Analysis}
\label{app:more_sensitivity}
Beyond the hyperparameter studies in Sec.~\ref{exp: rq4} (Table~\ref{tab:sensitivity}), we examine the effect of graph structural
properties on performance gains.

\paragraph{Sensitivity to graph homophily.}
\label{app:homophily}
RHEA relies on neighbor agreement to estimate reliability, and thus its effectiveness depends on attribute homophily. Across
benchmarks, edge homophily ranges from $0.68$ (Grocery) to $0.96$ (RedditS). As shown in Table~\ref{tab:robust_nmi}, RHEA
consistently improves over DGF under missing data settings across this range.

Performance gains decrease smoothly as homophily decreases, rather than exhibiting abrupt degradation, indicating that the method
remains stable but gradually approaches the behavior of non-reliability-aware baselines in heterophilous regimes
(see Appendix~\ref{app:limitations}).

\section{More Interpretability Investigation}
\label{app:more_interp}
\begin{figure}[t]
\centering
\includegraphics[width=0.7\columnwidth]{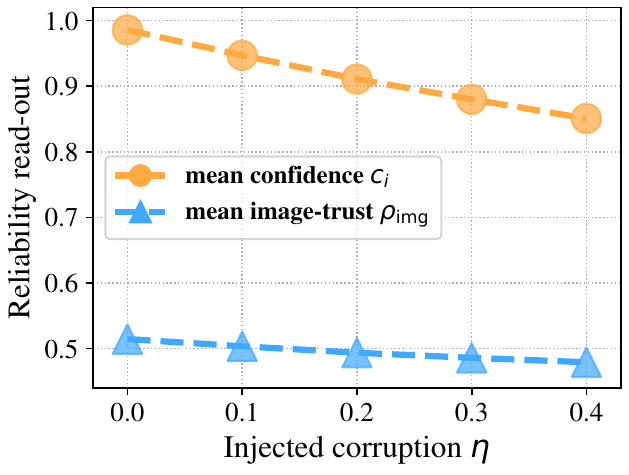}
\caption{Reconstruction confidence vs.\ corruption (RedditS). Mean confidence $c_i$ and image trust $\rho_{\mathrm{img}}$ versus corruption level $\eta$.}
\label{fig:conf_eta}
\end{figure}

We further analyze the behavior of the learned reliability signal. As presented in Fig.~\ref{fig:conf_eta}, the node-wise confidence score $c_i$ reflects agreement among donor neighbors and thus responds to input corruption even when all
modalities are present. On RedditS, the average confidence decreases monotonically from $0.986$ at $\eta=0$ to $0.851$ at
$\eta=0.4$, while image-trust $\rho_{\mathrm{img}}$ exhibits a consistent trend.

\section{Efficiency and Complexity Analysis}
\label{app:complexity}

\paragraph{Time complexity.} Modality-specific encoding and graph propagation rely on sparse message passing over $\mathcal{G}=(\mathcal{V},\mathcal{E})$, yielding $\mathcal{O}(|\mathcal{E}| d)$ per iteration, where $d$ is the embedding dimension. Reliability-aware reconstruction and fusion introduce only local neighborhood aggregation and remain bounded by the same sparsity pattern. 

The entropic optimal transport solver requires $T$ Sinkhorn iterations over an $N \times K$ cost matrix, resulting in $\mathcal{O}(NKT)$ complexity, and the inter-node repulsion term $\Omega_{\mathrm{cm}}$ (Eq.~\eqref{eq:omega} below) contributes an additional dense $\mathcal{O}(N^2 d)$ computation. Therefore, the overall per-iteration time complexity is
\[
\mathcal{O}(|\mathcal{E}| d + NKT + N^2 d),
\]
whose quadratic cross-modal term is shared with prior cross-modal graph clustering frameworks such as DGF, while all graph and transport components remain sparse or linear in $N$.

\paragraph{Memory complexity and the $O(N^2)$ term.} The graph-based components scale linearly with the number of edges and feature dimension. The only super-linear term arises from the inter-node margin repulsion $\Omega_{\mathrm{cm}}$ in $\mathcal{L}_{\mathrm{mod}}$ (Eq.~\eqref{eq:lmod}), which is defined as:
\begin{equation}
\Omega_{\mathrm{cm}} \;=\; \frac{1}{N^2}\sum_{i}\sum_{j\neq i}
\big[\, m_0 + \langle \bh_i,\bz^{(\mathrm{txt})}_j\rangle - \langle \bh_i,\bz^{(\mathrm{txt})}_i\rangle \,\big]_{+},
\label{eq:omega}
\end{equation}
where $m_0$ is the repulsion margin and $[\cdot]_{+}=\max(0,\cdot)$.

This term induces a dense $N \times N$ similarity computation $\bh\,\bz^{(\mathrm{txt})\top}$, leading to an $\mathcal{O}(N^2)$ memory requirement. All other operations—including reliability modeling, reconstruction, message passing, and Sinkhorn updates—remain sparse and scale at most linearly with graph size. Therefore, the overall memory complexity is dominated by this quadratic term, consistent with prior cross-modal graph clustering frameworks.
\section{Additional Experimental Details}
\label{app:details}

\subsection{Training Protocol}
\label{app:protocol}

Unless otherwise specified, all reported results are presented as mean$\pm$standard deviation over five random
seeds. For each seed, all methods share the same data split and, for robustness experiments, the same realization
of missing-modality or corruption patterns. Perturbations are applied consistently to either image or text
modalities under identical settings. All models are trained for 120 epochs following the same protocol. Baseline
implementations use the official released code with the recommended hyperparameter settings, while ablated variants
inherit the same configuration with only the corresponding component removed.

\paragraph{Perturbation model.}
Both perturbations are injected into the raw features of the target modality before encoding, and both ratios
are node fractions rather than noise amplitudes. Corruption with ratio $\eta$ selects $\lfloor \eta N \rfloor$
nodes uniformly at random and replaces each selected node's feature vector with a random Gaussian vector rescaled
to the $\ell_2$ norm of the original feature, so the content of the modality is destroyed while its scale is
preserved. Missingness with ratio $m$ selects $\lfloor m N \rfloor$ nodes uniformly at random and sets the target
modality's feature vector to zero. Node subsets are sampled with independent random streams per regime and per
modality, and for a given seed every method receives the identical node mask and noise realization.

\subsection{Model Configuration}
\label{app:hyper}

Each node is associated with text and image features extracted from frozen pretrained encoders. We use RoBERTa for
text and CLIP-ViT-L/14 for images, producing 768-dimensional representations for all MAGB datasets. The DMGC
Amazon dataset uses its provided 1433-dimensional node attributes. Unless otherwise stated, each modality
is encoded by a linear projection followed by simplified graph propagation.

The default hyperparameters are shared across all datasets. Specifically, the reconstruction threshold is
$\gamma=0.45$, the reliability temperature is $\tau=1$, the contrastive temperature is $\tau_c=0.5$, and the
entropic regularization strength in optimal transport is $\epsilon=0.05$ with 20 Sinkhorn iterations.
The remaining hyperparameters are likewise shared: the loss weights are $\lambda_{\mathrm{ncrc}}=1$,
$\lambda_{\mathrm{nbr}}=0.5$, and $\lambda_{\mathrm{mod}}=0.1$; the assignment temperature is $\tau_a=1$; the
consensus propagation depth is $L=2$; the sharpening temperature is $T_s=0.5$; the confidence decay coefficient
is $\beta=3$; and the repulsion margin is $m_0=0.1$. A
sensitivity analysis of the main hyperparameters is reported in Sec.~\ref{exp: rq4}.

\paragraph{Optimization.}
All models are optimized using Adam with learning rate $10^{-3}$ and weight decay $10^{-5}$ for 120 epochs under
full-batch training. The embedding dimension is fixed to $64$, and graph propagation uses 10 layers throughout all
experiments.

\paragraph{Reconstruction confidence.}
The reconstruction confidence combines neighborhood consistency and neighborhood support. Specifically, it
increases when neighboring donor embeddings exhibit high agreement and sufficient coverage, and decreases when
neighbor agreement is weak or only a few reliable donors are available. Nodes without reliable neighbors fall back
to their original modality representation, ensuring stable behavior across different graph structures.

\paragraph{Initialization and normalization.}
All modality-specific and fused embeddings are $\ell_2$-normalized before reliability estimation, reconstruction,
and optimal transport. Cluster centers are initialized by $k$-means on the fused embeddings from the first epoch
and optimized jointly during training.

\subsection{Stabilizing the Reliability Field}
\label{app:ema}

\begin{figure}[t]
\centering
\includegraphics[width=0.75\columnwidth]{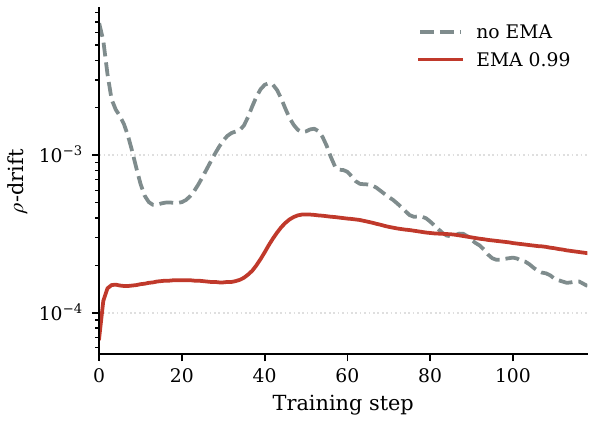}
\caption{Stability of the reliability field. Per-step variation of the reliability estimate $\rho$ with and without EMA on RedditS under missing-modality ratio $m=0.5$.}
\label{fig:ema}
\end{figure}

The reliability field is updated using an exponential moving average (EMA) with momentum 0.99 to improve training
stability. As illustrated in Fig.~\ref{fig:ema}, EMA substantially reduces fluctuations of the reliability
estimates across training while preserving clustering performance. We therefore adopt EMA as the default setting in
all experiments.

\subsection{Experimental Environment}
\label{app:environment}

All experiments are conducted on a single NVIDIA RTX PRO 6000 (Blackwell Server Edition, 96\,GB) GPU using PyTorch 2.10 with CUDA 12.8.
\section{Dataset Details}
\label{app:datasets}

We evaluate on four public multimodal attributed graphs; every node carries two modalities (text and image),
encoded by frozen pretrained models. Table~\ref{tab:datasets} lists their statistics, with textual descriptions
below.

\begin{table}[t]
\centering
\caption{Statistics of the four evaluation datasets. All datasets are two-modality (text${+}$image); the
feature dimensions are those of the frozen pretrained encoders.}
\label{tab:datasets}
\setlength{\tabcolsep}{4pt}
\renewcommand{\arraystretch}{1.05}
\resizebox{\columnwidth}{!}{%
\begin{tabular}{l ccccl}
\toprule[1.0pt]
\textbf{Dataset} & \textbf{\# Nodes} & \textbf{\# Clusters} & \textbf{Text dim} & \textbf{Image dim} & \textbf{Domain} \\
\midrule[0.1pt]
RedditS & 15,894 & 20 & 768 & 768 & Social \\
Toys    & 20,695 & 18 & 768 & 768 & E-commerce \\
Grocery & 17,074 & 20 & 768 & 768 & E-commerce \\
Amazon  & 6,158  & 3  & 1,433 & 1,433 & Co-purchase \\
\bottomrule[1.0pt]
\end{tabular}}
\end{table}

\paragraph{RedditS.}~\cite{RedditS} A social-network graph from the MAGB benchmark, sourced from RedCaps.
Each node is a post carrying its text (title and body) and its accompanying image; edges link related posts, and
the $20$ clusters correspond to their communities. Text and image features are the $768$-dimensional RoBERTa and
CLIP-ViT-L/14 embeddings, respectively.

\paragraph{Toys.}~\cite{Amazon2018} As a core component of MAGB, the Toys dataset comprises roughly $42$k
products (of which $20{,}695$ form the graph). Age suitability, safety information, materials, and functional
descriptions are encoded alongside visual features, and relations among products, including series membership and
accessory compatibility, form the graph edges. It spans $18$ product categories.

\paragraph{Grocery.}~\cite{Amazon2018} An Amazon-derived grocery co-purchase graph from MAGB with $17{,}074$
nodes and $20$ categories. Each product carries a textual description and a product image, and edges connect
co-purchased or related items. Text and image features follow the same $768$-dimensional RoBERTa / CLIP encoding as
the other MAGB graphs.

\paragraph{Amazon.}~\cite{guo2025dmgc} The multimodal co-purchase graph used by DMGC, with $6{,}158$ nodes
and only $3$ clusters. Nodes carry $1{,}433$-dimensional text and image attributes, and the graph combines
item--item and item--user relations. Its small cluster count and highly discriminative attributes make even simple
attribute-only baselines competitive, which is why it is a useful stress test for reliability-guided fusion.

\section{Baseline Details}
\label{app:baselines}

We compare against nine baselines spanning four families. All are run from their official code under the identical
protocol of Appendix~\ref{app:protocol}.

\paragraph{KMeans} clusters nodes by Lloyd's algorithm on their concatenated text and image features, using
no graph structure. It is the classical attribute-only reference point.

\paragraph{DFCN}~\citep{tu2021dfcn} is a deep fusion clustering network that couples an autoencoder with a
graph autoencoder through a structure-and-attribute fusion module, and refines the fused representation with a
triplet self-supervision objective.

\paragraph{DMoN}~\citep{tsitsulin2023dmon} trains a graph neural network with a differentiable
spectral-modularity objective and a collapse regularizer, producing soft cluster assignments that directly optimize
graph community structure.

\paragraph{MVGRL}~\citep{hassani2020mvgrl} is a contrastive multi-view method that maximizes the mutual
information between node-level and graph-level representations across two structural views, the adjacency and a
graph-diffusion view.

\paragraph{S3GC}~\citep{devvrit2022s3gc} performs scalable self-supervised graph clustering: a graph encoder
is trained contrastively with random-walk-based positive pairs, so it scales to large graphs while remaining
label-free.

\paragraph{COMPLETER}~\citep{lin2021completer} targets incomplete multi-view data through information-theoretic
dual prediction, maximizing cross-view mutual information while minimizing conditional entropy so that a missing
view can be recovered from the observed one.

\paragraph{APADC}~\citep{xu2022apadc} is a deep incomplete multi-view clustering method that is
imputation-free: it aligns the view distributions with a maximum-mean-discrepancy term and adaptively projects and
fuses whichever views are available for each sample.

\paragraph{DGF}~\citep{zheng2025dgf} jointly exploits graph topology and multiple modalities with a
dual-graph filtering backbone and a community-aware cross-modality contrastive objective. It is the strongest prior
method and our primary point of comparison.

\paragraph{DMGC}~\citep{guo2025dmgc} disentangles homophilous and heterophilous signals with a dual-frequency
(low- and high-pass) graph filter, combines intra- and cross-modal contrastive objectives, and refines the
assignment with a DEC clustering head.

\section{Metric Details}
\label{app:metrics}

We report four standard clustering metrics; all are higher-is-better. Let $Y=\{y_i\}_{i=1}^{N}$ be the
ground-truth labels and $C=\{c_i\}_{i=1}^{N}$ the predicted clusters over $N$ nodes. Because clustering is
permutation-invariant, ACC and F1 are computed after a best cluster-to-label matching obtained with the Hungarian
algorithm~\citep{kuhn1955hungarian}.

\paragraph{Normalized Mutual Information (NMI)}~\citep{strehl2002nmi} measures the shared information between
the predicted and ground-truth partitions, normalized by their entropies,
\begin{equation}
\mathrm{NMI}(Y,C) \;=\; \frac{2\,I(Y;C)}{H(Y)+H(C)} ,
\end{equation}
where $I(\cdot;\cdot)$ is mutual information and $H(\cdot)$ is entropy. It lies in $[0,1]$, with $1$ for a perfect
match.

\paragraph{Clustering Accuracy (ACC)} is the fraction of correctly assigned nodes under the best label
permutation $\pi$,
\begin{equation}
\mathrm{ACC} \;=\; \max_{\pi}\; \frac{1}{N}\sum_{i=1}^{N}\mathbb{1}\!\left(y_i = \pi(c_i)\right) ,
\end{equation}
where $\pi$ ranges over cluster-to-label mappings and $\mathbb{1}(\cdot)$ is the indicator function.

\paragraph{Adjusted Rand Index (ARI)}~\citep{hubert1985ari} counts agreeing node pairs, corrected for chance.
With $n_{ij}$ the number of nodes in ground-truth cluster $i$ and predicted cluster $j$, and $a_i,b_j$ the
corresponding row/column sums,
\begin{equation}
\resizebox{\columnwidth}{!}{$\displaystyle
\mathrm{ARI} \;=\; \frac{\sum_{ij}\binom{n_{ij}}{2} - \big[\sum_i\binom{a_i}{2}\sum_j\binom{b_j}{2}\big]\big/\binom{N}{2}}
{\tfrac{1}{2}\big[\sum_i\binom{a_i}{2}+\sum_j\binom{b_j}{2}\big] - \big[\sum_i\binom{a_i}{2}\sum_j\binom{b_j}{2}\big]\big/\binom{N}{2}}\,.
$}
\end{equation}

\paragraph{F1} is the macro-averaged harmonic mean of precision (P) and recall (R) over the matched clusters,
$\mathrm{F1} = 2\,\mathrm{P}\,\mathrm{R}/(\mathrm{P}+\mathrm{R})$, which is robust to class imbalance.

\section{Limitations and Broader Impact}
\label{app:limitations}

\paragraph{Limitations.}
RHEA assumes that local neighborhood information provides useful cues for estimating modality reliability and is therefore best suited to graphs with informative local structure. As neighborhood agreement becomes less informative, the benefit of reliability-aware reconstruction may diminish accordingly. In addition, RHEA inherits the dense cross-modal contrastive objective from DGF, whose $O(N^2)$ memory complexity (Appendix~\ref{app:complexity}) may limit scalability to very large graphs. Exploring reliability estimation under more challenging graph structures and improving scalability are promising directions for future work.

\paragraph{Broader Impacts.}
Multimodal attributed graphs are widely used in applications such as recommendation systems, social networks, and
citation analysis, where data quality is often heterogeneous across modalities. RHEA provides a label-free
mechanism to estimate modality reliability and improve robustness under imperfect data conditions, which may help
reduce sensitivity to noise in downstream analysis. As with other unsupervised representation learning methods,
care should be taken when applying the resulting embeddings or clusters in high-stakes decision scenarios.

% \newpage
% \def\isChecklistMainFile{}
% \input{ReproducibilityChecklist.tex}

\end{document}